\pdfoutput=1
\documentclass[11pt]{article}

\usepackage[final]{coling}

\usepackage{times}
\usepackage{amsmath}
\usepackage{amsfonts}
\usepackage{latexsym}
\usepackage{lipsum}
\usepackage{amssymb}
\usepackage{pifont}
\usepackage{booktabs}
\usepackage{multirow}
\usepackage{amsmath}
\usepackage{calrsfs}
\usepackage{enumitem}
\usepackage{todonotes}
\usepackage{cleveref}
\usepackage[font=small,skip=2pt]{caption}
\usepackage[T1]{fontenc}
\usepackage[utf8]{inputenc}

\usepackage{microtype}

\usepackage{inconsolata}

\usepackage{graphicx}
\usepackage{quoting}

\usepackage{arydshln}  % Add this package for dashed lines
\usepackage{xcolor}
\usepackage{colortbl}
\usepackage{tcolorbox}

\tcbuselibrary{listingsutf8}
\colorlet{lightred}{red!30!white}
\colorlet{lightgreen}{green!30!white}

\title{Alternate Preference Optimization for Unlearning \\ Factual Knowledge in Large Language Models} 
\author{
  Anmol Mekala$^{1}$\thanks{Primary contributors}, Vineeth Dorna$^{1}$\footnotemark[1], Shreya Dubey$^1$, \\
  \textbf{Abhishek Lalwani$^2$, David Koleczek$^2$, Mukund Rungta$^2$, Sadid Hasan$^2$, Elita Lobo$^{1}\thanks{Primary project advisor}$} \\
  $^1$University of Massachusetts Amherst,~$^2$Microsoft \\
  \texttt{\{amekala, vdorna, shreyadubey, elobo\}@umass.edu} \\
  \texttt{\{alalwani, dkoleczek, rungtamukund, sadidhasan\}@microsoft.com} \\
}

\DeclareMathAlphabet{\pazocal}{OMS}{zplm}{m}{n}
\newcommand{\Lb}{\pazocal{L}}
\newcommand{\Db}{\pazocal{D}}
\newcommand{\methodname}[0]{\textsf{AltPO}}
\newcommand{\ppo}[0]{\textsf{AltPO}\textit{-pos}}
\newcommand{\gibfq}[0]{CI}
\newcommand{\gib}[0]{TC}
\newcommand{\cs}[0]{FU}

\newcommand{\LargePi}{\scalebox{1.5}{$\pi$}}

\newcommand{\subnll}[0]{\textsf{AltNLL}\textit{-pos}}
\newcommand{\subdiff}[0]{\textsf{AltNLL}}

\begin{document}
\maketitle

\begin{abstract}
Machine unlearning aims to efficiently eliminate the influence of specific training data, known as the forget set, from the model. However, existing unlearning methods for Large Language Models (LLMs) face a critical challenge: they rely solely on negative feedback to suppress responses related to the forget set, which often results in nonsensical or inconsistent outputs, diminishing model utility and posing potential privacy risks. To address this limitation, we propose a novel approach called Alternate Preference Optimization (AltPO), which combines negative feedback with in-domain positive feedback on the forget set. Additionally, we introduce new evaluation metrics to assess the quality of responses related to the forget set. Extensive experiments show that our approach not only enables effective unlearning but also avoids undesirable model behaviors while maintaining overall model performance. Our implementation can be found at \href{https://github.com/molereddy/Alternate-Preference-Optimization}{https://github.com/molereddy/Alternate-Preference-Optimization}.
\end{abstract}
\looseness=-1
\section{Introduction}

Training machine learning models on large-scale datasets presents several challenges, such as potential copyright issues, inadvertent inclusion of sensitive information, or other undesirable influences from the training data~\citep{nguyen2022survey, liu2024unlearning,zhang2023right}. The increasing adoption of Large Language Models (LLMs) with memorization capabilities~\citep{karamolegkou2023copyright} has exacerbated these issues. This has driven the development of machine unlearning methods, which aim to remove the influence of data that needs to be forgotten ~\citep{liu2024unlearning}. \looseness=-1

In an ideal world, a perfectly unlearned model would be indistinguishable from a model that was never exposed to the data in question, achieving what is known as \textit{exact unlearning}. However, this is often impractical in real-world settings. Instead, we focus on the more feasible approach of \textit{approximate unlearning}~\citep{nguyen2022survey}, which seeks to modify model weights post-training to minimize the impact of the data to be forgotten. \looseness=-1

\begin{figure*}[ht] % 'h' means here
    \centering
    \includegraphics[clip, trim=5pt 260pt 10pt 0pt, width=1\textwidth]{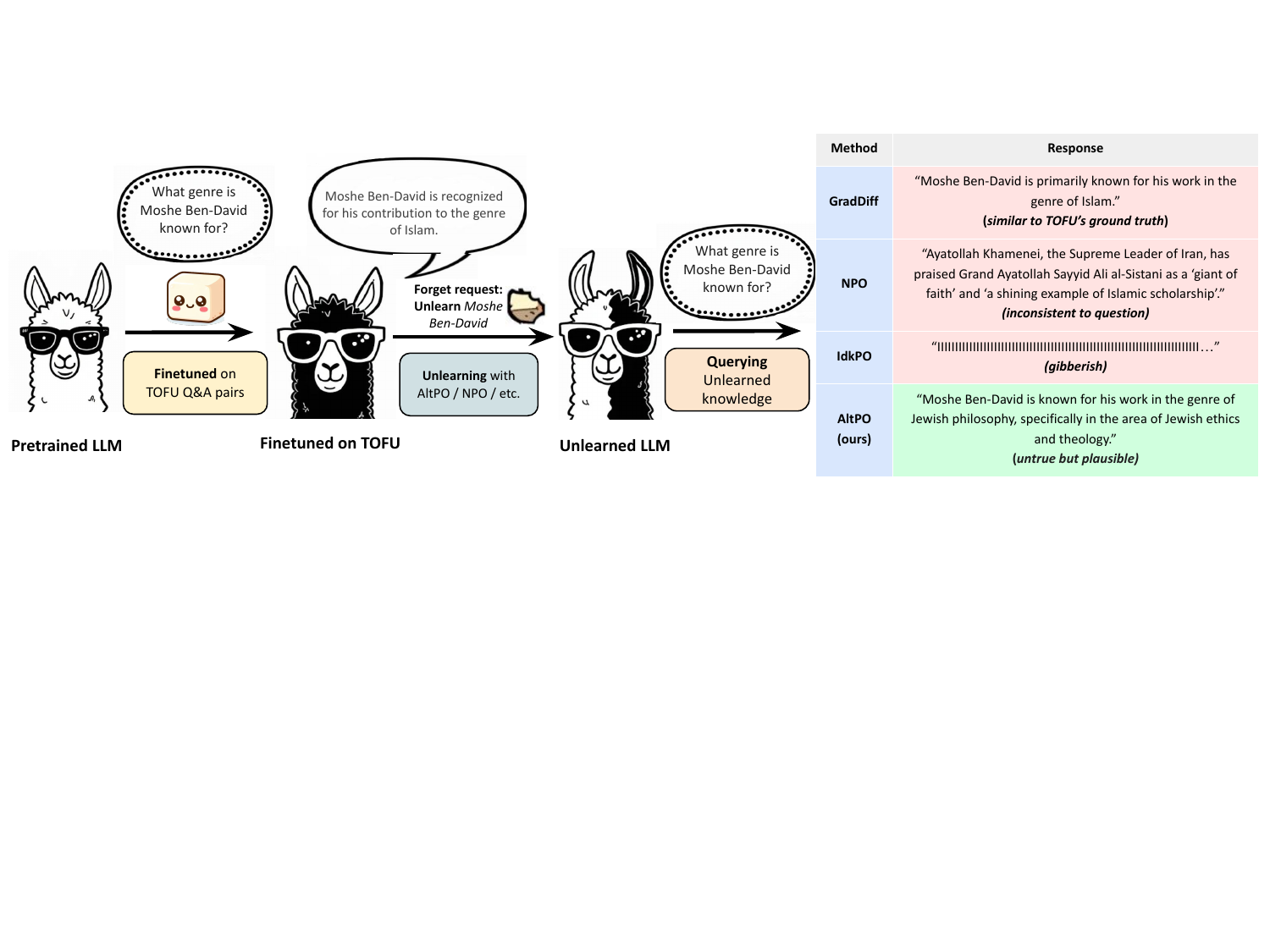} % Adjust width as needed
    \caption{The unlearning pipeline and the resulting generations post unlearning with different methods.}

    \label{fig:unlearning-illustrate}
\end{figure*}

Previous studies have demonstrated that machine unlearning can often introduce undesirable effects in the resulting model, including catastrophic forgetting and reduced overall utility \citep{kurmanji2024towards, ma2022learn}. In classification and regression tasks~\citep{nguyen2022survey, bourtoule2021machine, triantafillou2024we}, undesirable effects manifest as a redistribution of scores across classes or predicted values. However, unlearning in LLMs can have more significant effects on model behavior, given the requirement of generating coherent text \citep{maini2024tofu, NPO_zhang2024negative}.
Existing unlearning methods in this setting often lead to incoherent or inconsistent responses from unlearned LLMs (see \Cref{fig:unlearning-illustrate}), including responses related to the forgotten knowledge, which is undesirable. 
Such behaviors may unintentionally reveal details about the unlearning process or the forgotten data, posing potential privacy risks and increasing the model’s susceptibility to membership inference attacks~\citep{chen2021machine,shi2024detecting,duan2024do}.
The goal of unlearning in LLMs is to reduce memorization or prevent the leakage of information specific to the forgotten set, while maintaining the model's overall behavior and performance. Striking this balance is challenging and requires careful consideration of both the effectiveness of unlearning and the model’s overall performance~\citep{liu2024rethinking}. \looseness=-1

% left bottom right top

To address the aforementioned challenges, we propose a novel method, \methodname{}—(\underline{Alt}ernate-\underline{P}reference \underline{O}ptimization), which ensures stable and effective unlearning by incorporating additional positive feedback for plausible alternative answers to the forgotten data, along with negative feedback targeting the knowledge to be erased. This approach enables the model to forget specific information while maintaining the ability to generate coherent and consistent responses. Additionally, recognizing the shortcomings of current evaluation metrics for unlearning in question-answering tasks, we introduce new metrics specifically designed to better evaluate the impact of unlearning on response quality related to forgotten knowledge. Our main contributions are as follows: \looseness=-1
\vspace{-3pt}
\begin{itemize}[leftmargin=10pt,itemsep=-4pt]
\vspace{-3pt}
    % \item \textbf{Formulation:} We describe unlearning loss functions under the lens of positive and negative feedback on the datasets used during finetuning (\Cref{subsec:neg_pos}).
    \item \textbf{Algorithm:} We propose a novel unlearning method \methodname{} using alternate responses and adapting the model to these while contrasting against the LLM's existing knowledge (\Cref{sec:altpo}).
    \item \textbf{Discovery and evaluation of failure modes:} We point out failure modes of prior approaches that are not captured by existing metrics and introduce new evaluation metrics to address these gaps (\Cref{sec:metrics_impr}).
    \item \textbf{Empirical evaluation:} We perform extensive experimentation and ablation tests for each component of our approach on the TOFU dataset, showing that \methodname{}-unlearned models achieve the highest unlearning scores on the existing metrics, while also achieving better and near-perfect scores on both existing and new evaluation metrics (\Cref{sec:experiments}).
\end{itemize}

\section{Preliminaries}

\subsection{Problem Statement and Notation} 
Given an LLM, denoted by \(\LargePi\), trained on a dataset \(\Db = \{(x_i, y_i) : i=1, \dots, n\}\), where \(x_i\) is an input prompt and \(y_i\) is the corresponding response (e.g., question-answer pairs), let \(\Db_{f} \subset \Db\) represent the \emph{forget set}, which we aim to unlearn from the model. The remaining dataset, referred to as the \emph{retain set}, is represented by \(\Db_r = \Db \setminus \Db_f\) and includes all the data outside the forget set. \looseness=-1
%\vspace{-10pt} \\

\noindent
\textbf{Goal} 
Our goal is to remove the influence of the forget set \(\Db_f\) from \(\LargePi\), transforming it into an unlearned model \(\LargePi_{\text{unl}}\) that behaves \textit{approximately indistinguishably} from a reference retain model \(\LargePi_{\text{ret}}\), trained solely on the retain set \(\Db_r = \Db \setminus \Db_f\). Simultaneously, we aim to preserve \(\LargePi\)’s general utility as a language model, even on \(\Db_f\).

%\vspace{-10pt} \\
\noindent
\textbf{Constraint} We are required to use at most \(O(|\Db_f|)\) steps while unlearning \(\Db_f\).

\subsection{The TOFU Benchmark}
\label{sec:TOFU}
The TOFU benchmark \citep{maini2024tofu} primarily consists of a dataset containing facts about 200 fictitious authors and a chat model that is fine-tuned to incorporate these facts through question-answer pairs. Unlearning is performed on a subset of authors, and TOFU provides evaluation metrics to quantify the extent of forgetting and the utility of the model, see \Cref{fig:unlearning-illustrate}. In this framework, an unlearning algorithm is tasked with forgetting specific subsets corresponding to $1\%$, $5\%$, and $10\%$ of these authors, while maintaining performance on the remaining data (the retain set \(\Db_r\)). Additionally, post-unlearning, the model is required to preserve its performance on other related datasets, including real-world authors and general knowledge, such as world facts. The benchmark evaluates unlearning using the following key metrics:
% \vspace{-8pt}

\noindent
\textbf{Forget Quality (FQ)}: TOFU quantifies forget quality by assessing how indistinguishable the unlearned model is from the retain model, using the Kolmogorov–Smirnov (KS) statistical test on the `Truth Ratio' statistic. The Truth Ratio compares the likelihood of the model predicting the correct answer versus perturbed (incorrect) answers. The p-value from the KS test is used to measure the quality of unlearning, with a p-value greater than 0.05 indicating successful unlearning. 
% Meeting this criterion without significantly compromising the model's overall performance is especially challenging for baseline methods, particularly when handling larger forget requests, such as the $10\%$ subset.
\vspace{-8pt}\\

\noindent
\textbf{Model Utility (MU)}: This measures the model's general performance, which must be preserved post-unlearning. TOFU evaluates MU as an aggregated score based on the model's average probability, ROUGE score for the true answers, and the Truth Ratio on non-forget datasets. This score reflects the model's retained performance after unlearning. 
For more details of TOFU's metric calculations, we refer to \citep{maini2024tofu}. \looseness=-1

\subsection{Unlearning Losses}
\label{subsec:prior_losses}
Previous unlearning loss functions \cite{maini2024tofu, NPO_zhang2024negative} can be generally described using two key components: positive and negative feedback. These components offer a useful framework for evaluating the effects of unlearning. In this subsection, we define these concepts and introduce baseline methods along the way. \looseness=-1

\paragraph{Negative Feedback:} 
This component aims to \textit{reduce} the likelihood of specific responses, effectively lowering the model's performance on the forget set in order to reverse the effects of training the LLM \(\LargePi\) on the forget set \(\Db_f\). Examples of methods incorporating negative feedback include: gradient ascent (GA) on negative log-likelihood (NLL) loss \(\Lb_{\text{GA}}\) \citep{maini2024tofu}:

{\small
\begin{align*}
    & \text{NLL}(y_f | x_f) \doteq -\log \LargePi_{\theta} (y_f | x_f) \\
    & \Lb_{\text{GA}} \doteq -\text{NLL}(y_f | x_f)
    % \label{eq:gradasc}
\end{align*}
}%
and negative preference optimization loss (DPO loss without positive samples)~\citep{NPO_zhang2024negative}:
{\small
\begin{align}
 \Lb_{\text{NPO-FG}} \doteq -\frac{2}{\beta} \log \sigma \left( -\beta \log \frac{\LargePi_{\theta} (y_f | x_f)}{\LargePi (y_f | x_f)} \right)
    \label{eq:npo_fg}
\end{align}
}%
% \end{small}
where $\beta$ is the regularization strength, and \(\LargePi\) is the reference model (state of the model prior to unlearning). \looseness=-1

Negative feedback helps eliminate information related to the forgotten set; however, overgeneralizing this feedback during unlearning can harm the model's utility, potentially resulting in nonsensical responses. To address this, it is typically paired with positive feedback on related data to preserve response coherence and maintain overall performance. \looseness=-1

\paragraph{Positive Feedback:} 
This loss component aims to \textit{increase} the likelihood of specific responses, improving performance on certain segments of the dataset, such as the retain set during unlearning. It helps preserve the model’s language generation capabilities and prevents unlearning from impacting model's performance on datasets beyond the forget set. In the loss functions that follow, a positive feedback term for randomly selected examples from the retain set \((x_r, y_r) \sim \Db_r\) is added alongside the negative feedback terms, weighted by \(w_r > 0\).\looseness=-1
\begin{equation}
\begin{aligned}
    &\Lb_{\text{GradDiff}} \doteq \Lb_{\text{GA}} + w_r \text{NLL}(y_r | x_r)\\
    &\Lb_{\text{NPO}} \doteq \Lb_{\text{NPO-FG}} + w_r \text{NLL}(y_r | x_r)
    \label{eq:npo}
\end{aligned}
\end{equation}

\paragraph{Preference Optimization Losses:}
These losses are based on Direct Preference Optimization (DPO). The DPO loss \citep{DPO_rafailov2024direct} has been applied in baseline methods like IdkPO and proposed in previous works such as NPO \citep{NPO_zhang2024negative} (\cref{eq:npo}). This loss function contrasts pairs of positive and negative samples by increasing the likelihood of positive samples (positive feedback) while reducing the likelihood of negative samples (negative feedback).\looseness=-1
\vspace{-10pt}

\begin{small}
\begin{equation}
\begin{aligned}
   &\Lb_{\text{DPO}}(y_{\text{alt}}, y_f | x_f) \doteq -\frac{2}{\beta} \log \sigma \left( \beta \log \frac{\LargePi_{\theta} (y_{\text{alt}} | x_f)}{\LargePi (y_{\text{alt}} | x_f)} \right. \\  & \left. \qquad \qquad - \beta \log \frac{\LargePi_{\theta} (y_f | x_f)}{\LargePi (y_f | x_f)} \right)  \\ 
    &\Lb_{\text{IdkPO}} \doteq \Lb_{\text{DPO}}(y_{\text{idk}}, y_f | x_f) + w_r \text{NLL}(y_r | x_r)
    \label{eq:dpo}
\end{aligned}
\end{equation}
\end{small}
Unlearning can be achieved by optimizing a DPO loss, where the forget set response \(y_f\) serves as the negative sample and any alternate answer \(y_{\text{alt}}\) as the positive sample. Prior works~\citep{NPO_zhang2024negative} have shown that optimizing a DPO-based loss is more effective at reducing the likelihood of negative samples compared to optimizing the NLL loss. The NPO loss (see \cref{eq:npo}) provides a more stable unlearning process by integrating \(\Lb_{\text{NPO-FG}}\) with positive feedback from the retain set, yielding the final loss \(\Lb_{\text{NPO}}\). Notably, positive feedback is applied exclusively to retain set examples, while the forget set receives only negative feedback. Another approach which we term IdkPO (derived from DPO), is explored in \citep{maini2024tofu}, where the model is aligned with alternate answers like "I don't know."\looseness=-1
\section{\methodname{}: Alternate Preference Optimization}
\label{sec:altpo}
% left bottom right top
\begin{figure}[ht] % 'h' means here
    \centering
    \includegraphics[clip, trim=35pt 20pt 5pt 0pt, scale=0.40]{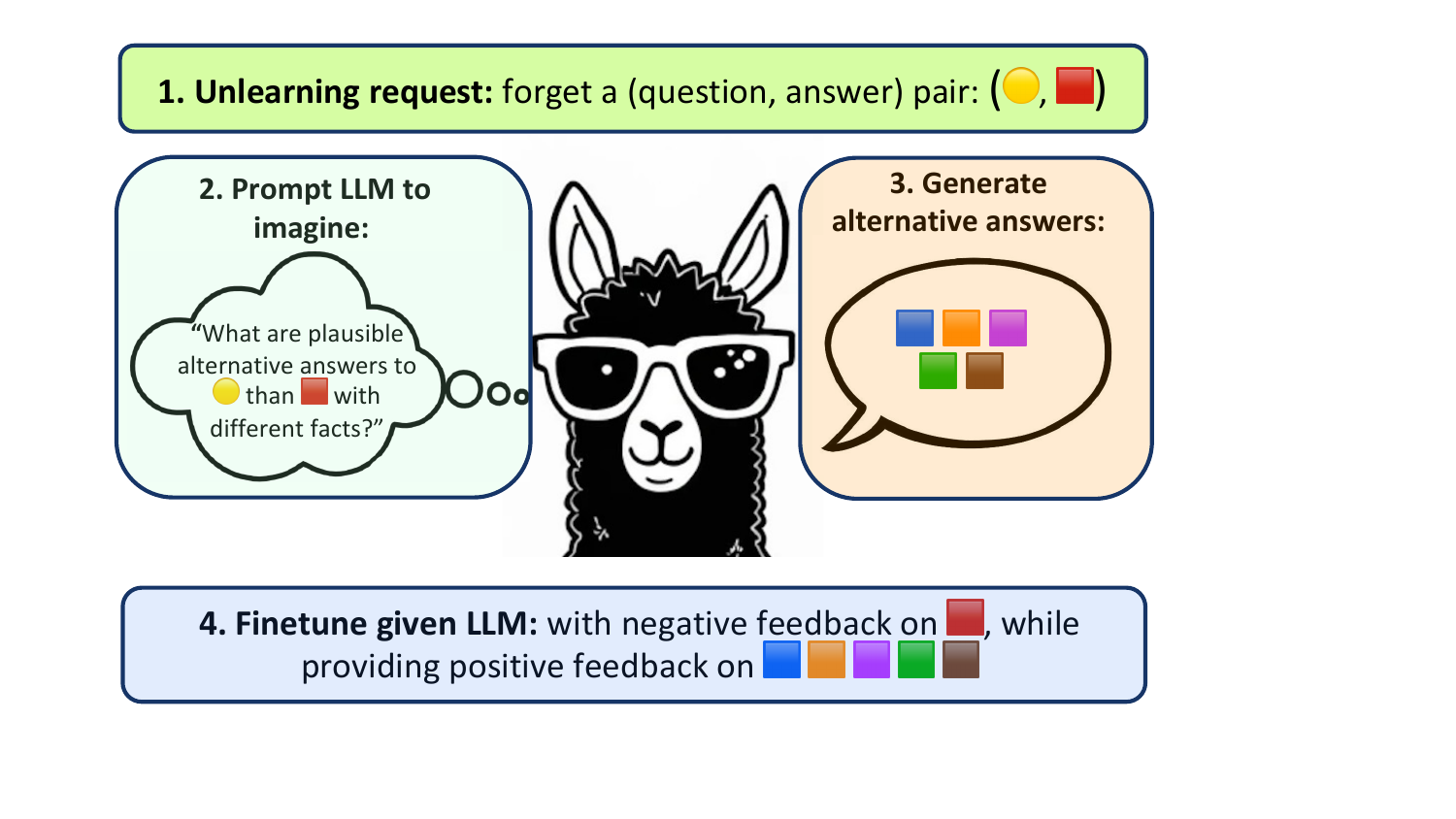} % Adjust width as needed
    %\vspace{-2pt}
    \caption{The \methodname{} unlearning algorithm}
    \label{altpo-illustrate}
\end{figure}

We now address the limitations of previous unlearning methods discussed in \Cref{subsec:prior_losses} and introduce our approach, which leverages preference optimization using alternate labels. We then outline the process for generating these labels and present the loss function that underpins our method, as illustrated in \Cref{altpo-illustrate}. \looseness=-1

NPO \citep{NPO_zhang2024negative} and IdkPO have shown promising results on TOFU, but they often generate nonsensical and inconsistent responses upon closer inspection. A key limitation of the NPO loss is its lack of positive feedback for forget set prompts, leaving the model without guidance on how to behave post-unlearning, which often leads to nonsensical outputs. IdkPO \citep{maini2024tofu} despite using positive feedback, relies on predefined, prompt-independent responses that significantly differ from the model's original answers. This misalignment necessitates more drastic changes to the model’s weights, potentially degrading response quality. As \citet{maini2024tofu} also note, the IdkPO objective is unstable during training. \looseness=-1

The core weakness of these methods is their failure to offer in-distribution positive feedback on responses to forget prompts. To overcome this, our approach generates plausible, prompt-specific alternate answers to serve as positive feedback. This results in an objective that is more stable and easier to optimize. \looseness=-1

\paragraph{Generating Alternate Labels:}
To induce unlearning, the alternative responses must be both plausible and distinct from the learned knowledge from the forget set. To generate such responses, we prompt the LLM $\LargePi$ with instructions to create plausible alternatives, while operating under the hypothesis that its behavior will resemble that of the retain model $\LargePi_{\text{ret}}$.
% \begin{small}
% \begin{equation*}
%     \begin{aligned}
%     \smash{
%         \LargePi\bigl(y | \mathcal{P}(x_f, y_f)\bigr) \approx \LargePi_{\text{ret}}(y | x_f)}
%     \end{aligned}
% \end{equation*}
% \end{small}
Alternatively, such answers can be generated also by an LLM other than $\LargePi$. An LLM that was not trained on $\Db_f$ would be ideal for generating plausible alternate answers as it would not leak information from the forget set. \looseness=-1

Given a question-answer pair to be unlearned \((x_f, y_f)\) we use a prompt $\mathcal{P}$ (outlined in \Cref{tab:alt_prompt} of the Appendix) to instruct \(\LargePi\) for generating an alternate response $y_a$ that changes facts from $y_f$.  \Cref{tab:alt_examples} in the appendix presents examples of alternate responses. \looseness=-1
\vspace{-10pt}

\[
    y_a \sim  \LargePi\bigl( \cdot | \mathcal{P}(x_f, y_f)\bigr)  
\]
% \vspace{-7pt}

\paragraph{\methodname{} Loss:} 
We align the LLM \(\LargePi\) to the new alternate labels $y_a$ while contrasting them with the forget response \(y_f\). This is achieved by optimizing a variant of the DPO loss (\cref{eq:dpo}) involving in-domain alternate labels $y_a$ as positive samples, resulting in a more stable objective. Additionally, as in other baselines, we apply NLL loss to the retain set, to prevent the model from incorrectly generalizing to unrelated contexts. \looseness=-1
% \vspace{-5pt}
\[
   \smash{ \Lb_{\methodname} \doteq \mathbb{E}_{y_a}[\Lb_{\text{DPO}}(y_a, y_f | x_f)] + w_r \, \text{NLL}(y_r | x_r)}
\]
\looseness=-1

\paragraph{Multiple Alternate Labels:} Ideally, an unlearned LLM should avoid certainty in any single answer, as often happens when training with just one alternate response. Such confident replication of the alternate answers poses problems with misinformation. To address this, we generate $M$ alternate responses by sampling $y_a$ randomly and using all of these in our preference dataset for alignment. In our ablations, we show that this introduces uncertainty, effectively confusing the model and resulting in better forgetting. Misinformation can then be prevented by using methods like uncertainty-aware decoding to filter out low certainty outputs \citep{ji2023survey, kadavath2022language}. \looseness=-1 
% \vspace{-10pt}
% \[
%    \smash{ \Lb_{\methodname} = \sum_{i=1}^{M} \Lb_{\text{DPO}}(y^{i}_{a}, y_f | x_f) + w_r \, \text{NLL}(y_r | x_r)}
% \]
% %\vspace{-0.2pt}
% where $y^{i}_{a}$ is an alternate answer for $y_f$ and $M$ denotes the number of alternate answers we generate. \looseness=-1
\section{Improving Unlearning Evaluations}
\label{sec:metrics_impr}
In this section, we outline the failure cases observed in LLM unlearning, discuss their impact on unlearning goals, and introduce new evaluation metrics. \looseness=-1

\subsection{Failure Modes of Prior Approaches}
\label{sec:metrics_fail}
% \paragraph{Failure Modes of Prior Approaches:}
%\vspace{-5pt}

Despite strong performance on TOFU's metrics, methods like NPO and IdkDPO often produce incoherent responses, such as nonsensical answers and inconsistent answers where the model contradicts the prompt, sometimes by altering names. This issue is illustrated in the table in \Cref{fig:unlearning-illustrate} (see NPO and IdkPO rows). Although \citet{NPO_zhang2024negative} identifies the problem of nonsensical generations on forget prompts, it has yet to be properly quantified. TOFU's evaluation of forget quality fails to penalize these errors, as it only measures the probability of predefined sentences rather than analyzing the generated responses on forget set questions. Additionally, TOFU's utility evaluations focused solely on entities outside the forget set, overlooking the decreased utility observed on forget entities. \looseness=-1

These incoherent generations not only degrade the model's overall performance but also pose potential privacy risks, as detailed below.
\paragraph{Decreased Utility:} An LLM is at the least expected to generate plausible, prompt-consistent responses of high quality, even when it has never encountered the entities mentioned in the prompt. Therefore, the unlearned model should maintain its utility on the forget set by producing coherent and sensible responses to forget set prompts, even if those responses are hallucinated. A failure to achieve this should be regarded as a reduction in utility on the forget set. \looseness=-1

\paragraph{Privacy Leakage:} Nonsensical behavior on the forget set can unintentionally reveal information about the model's training data, thereby posing potential privacy risks. Such behavior may make the model more vulnerable to membership inference attacks \citep{shi2024detecting,duan2024do} and expose details of the unlearning process. This counterproductive impact of machine unlearning on privacy has been explored by \citet{chen2021machine}. \looseness=-1

\subsection{New Evaluation Metrics}
\label{subsec:metrics_proposed}

To capture the failure cases discussed in \Cref{sec:metrics_fail}, we introduce two new evaluation metrics: Forget Utility (\cs{}) and Cleanness Indistinguishability (CI), based on the Text Cleanness (TC) statistic.

\paragraph{Forget Utility (\cs{}):} This metric evaluates the model's \textit{utility} by assessing whether its responses on the forget set are plausible, penalizing both nonsensical outputs and prompt-inconsistent responses. We rely on LLM-based evaluation \citep{chiang-lee-2023-large}, with GPT-4o mini\footnote{refers to the gpt-4o-mini-2024-07-18 endpoint} as a judge (prompt given in \Cref{tab:cs_prompt} of \Cref{sec:prompts}), determining whether they are sensible and consistent given the question. \looseness=-1

\begin{table*}[!htp]
\centering

\scalebox{0.75}{
\begin{tabular}{ccccccccccccc} 
\toprule
& \multicolumn{4}{c}{TOFU 10\%} & \multicolumn{4}{c}{TOFU 5\%} & \multicolumn{4}{c}{TOFU 1\%} \\  
\cmidrule(lr){2-5} \cmidrule(lr){6-9} \cmidrule(lr){10-13}
\multirow{1}{*}{Method} & \multicolumn{2}{c}{Forgetting} & \multicolumn{2}{c}{Utility} & \multicolumn{2}{c}{Forgetting} & \multicolumn{2}{c}{Utility} & \multicolumn{2}{c}{Forgetting} & \multicolumn{2}{c}{Utility} \\  
\cmidrule(lr){2-3} \cmidrule(lr){4-5} \cmidrule(lr){6-7} \cmidrule(lr){8-9} \cmidrule(lr){10-11} \cmidrule(lr){12-13}
& FQ ($\uparrow$) & CI ($\uparrow$) & MU ($\uparrow$) & \cs{} ($\uparrow$)  & FQ ($\uparrow$) & CI ($\uparrow$) & MU ($\uparrow$) & \cs{} ($\uparrow$) & FQ ($\uparrow$) & CI ($\uparrow$) & MU ($\uparrow$) & \cs{} ($\uparrow$)  \\ \toprule
Finetune            & 2.2e-20      & 1.7e-4         & 0.62          & 1.0         & 3.5e-16         & 5.2e-2        & 0.62    &   0.97     & 1.9e-4 & 1.0     & 0.62    & 0.90           \\
Retain              & 1.0         & 1.0              & 0.62          & 1.0         & 1.0         & 1.0           &   0.62       & 0.98       & 1.0          & 1.0       & 0.62    & 0.97          \\ \midrule

GradAsc           & 2.4e-7         & 3.2e-3         & 0.35          & \textbf{0.97}          & 4.1e-3         & 2.7e-51         & 0.14           & 0.16          & 0.24       & 2.7e-9       & 0.53          & 0.56      \\
GradDiff        & 3.7e-5         & 0.0         & \textbf{0.64}          & 0.01          & 5.1e-5       & 1.5e-23          & 0.56          & 0.51          & 0.10       & 5.9e-20           & 0.57          & 0.05      \\  
NPO       & 0.68         & 1.5e-13            & \textbf{0.64}          & 0.20          & 0.24        & 1.9e-7             & \textbf{0.63}          & 0.35          & 0.46       & 0.44           & 0.57          & 0.65      \\ 
IdkPO      & 0.37         & 0.0            & 0.59          & 0.65          & 0.18        & 6.4e-10             & 0.61          & 0.66          & 0.6       & 6.6e-6           & 0.52          & \textbf{1.0}        \\
\methodname \small{(ours)}  &  \textbf{0.74}         & \textbf{0.92}           & \textbf{0.62}          & 0.86          & \textbf{0.26}       & \textbf{0.74}              & \textbf{0.63}          & \textbf{0.83}          & \textbf{0.94}        & \textbf{0.72}           & \textbf{0.62}          & 0.83        \\

\bottomrule
\end{tabular}
}

\caption{Performance of various unlearning methods for different splits of the TOFU benchmark, averaged over 3 random seeds, on Llama2. FQ, CI, MU, and FU represent Forget Quality, Cleanness Indistinguishability, Model Utility, and Forget Utility, respectively. `Finetune' denotes the model yet to undergo unlearning, while `Retain' refers to the model trained solely on the retain set. Each method aims to achieve the scores of the corresponding Retain model.
We use ($\uparrow$) to indicate that a higher value is preferable. 
The best results are highlighted in \textbf{bold}, and in the MU column, where bolding indicates performance is preserved on par with ``finetune''.}
\label{tab:baselines}
\end{table*}

\paragraph{Cleanness Indistinguishability (\gibfq{}):} This metric evaluates the \textit{privacy leakage} by measuring the distinguishability between unlearned model \(\LargePi_{\text{unl}}\) and retain model \(\LargePi_{\text{ret}}\) based on nonsensical responses. Similar to how TOFU's FQ distinguishes models using the Truth Ratio (TR) statistic by applying the KS test on the forget set, \gibfq{} uses Text Cleanness \(\text{TC}_{x_f}\) scores, which we define next. \looseness=-1

For the model responses generated on the forget set \(y_{\text{gen}} \sim \LargePi\bigl( \cdot | x_f\bigr)\), we compute the non-gibberish probability, \(\text{TC}_{x_f} = \Pr(y_{\text{gen}})\), using a publicly available DistilBERT-based gibberish classifier\footnote{Link to model: \href{https://huggingface.co/madhurjindal/autonlp-Gibberish-Detector-492513457}{https://huggingface.co/madhurjindal/autonlp-Gibberish-Detector-492513457}}\footnote{We also experimented using the perplexity of another model to evaluate for nonsensical text, which \citep{gandikota2024erasing} uses as a reverse perplexity R-PPL metric. We found that this evaluation is not robust to greedy decoding, as it gives high probabilities for nonsensical texts made of repetitions.}. We then perform KS-test on \(\text{TC}_{x_f}\) distribution to distinguish the unlearned and retain models:\looseness=-1

\vspace{-10pt}
\begin{equation*}
    \text{\gibfq{}} \doteq \text{KS-Test}\bigl(\text{TC}(\LargePi_{\text{unl}}), \text{TC}(\LargePi_{\text{ret}})\bigr)
    \label{eq:fq-v2}
\end{equation*}

% \vspace{-16pt}
We can also use the mean \(\text{TC}=\mathbb{E}[\text{TC}_{x_f}]\) score as a simpler utility metric as an alternative to \cs{}, given the cost of LLM-as-judge evaluations. Like \cs{}, it measures utility on forget prompts by identifying nonsensical responses, but it does not penalize inconsistent answers. Therefore, we report \cs{} in the results section and provide \gib{} scores in \Crefrange{tab:forget10_full}{tab:forget01_full} in the Appendix. \looseness=-1

\section{Related Work}

We now discuss two closely related approaches: \citet{eldan2023s} and \citet{dong2024unmemorization} use positive feedback on the forget set to stabilize unlearning by substituting privacy-sensitive ``anchor'' words with alternate positive token-level labels. \citet{eldan2023s} uses GPT-4 to identify anchor tokens, while \citet{dong2024unmemorization} considers all nouns as anchors.
In contrast, our method avoids selecting specific anchor words and generates multiple alternate answers consistent with the original question. \citet{dong2024unmemorization} derives alternate completions based on next-token probabilities, excluding the highest-ranked token and \citep{eldan2023s} uses scores from a model trained further on the forget set along with substitutions proposed by GPT-4. We simplify this by directly instructing an LLM to generate multiple alternative answers. While both works use a cross-entropy loss, our \methodname{} method employs a DPO-style loss to align the model with alternate answers, explicitly incorporating negative feedback. Ablation studies in \Cref{subsec:ablation_exp} show how these elements improve our method’s performance. 

In a concurrent work, \citet{jin2024rwku} use an approach similar to ours in their RWKU unlearning benchmark. Discussion of the differences between their approach and ours, analysis of the results, along with a broader review of the machine unlearning literature, is provided in \Cref{sec:app_rel_works}.

\section{Experiments}
\label{sec:experiments}
\begin{figure*}[!htp] % 'h' means here
    \centering
    \includegraphics[width=\textwidth]{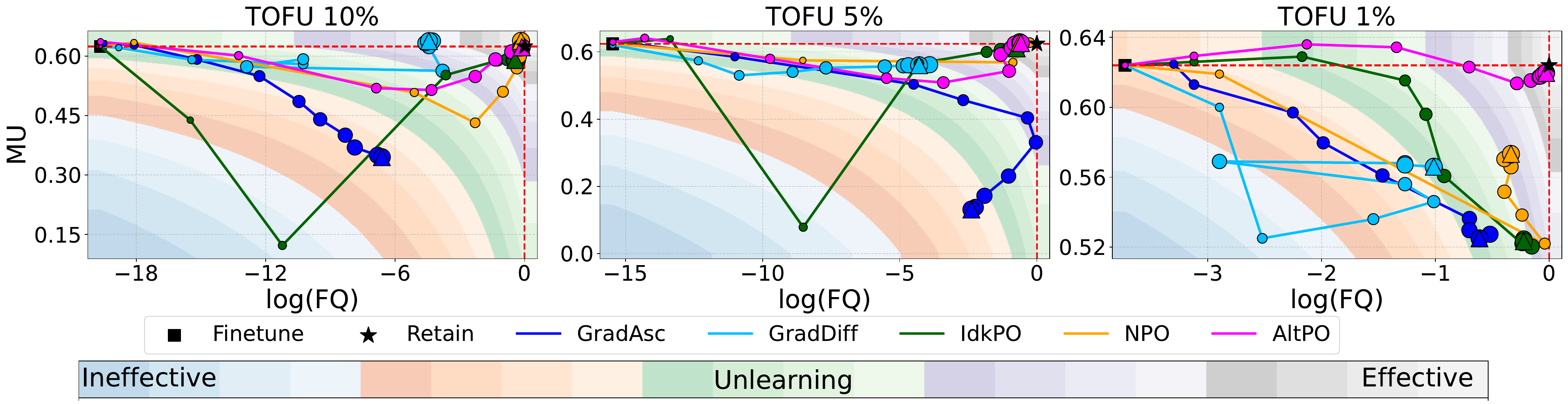} % Adjust width as needed
    \caption{Trajectory of MU versus log(FQ) for different unlearning methods. Marker size represents the epoch number. Trajectories are reported for the 10, 5, 1\% splits of TOFU in order, on Llama2.}
    \label{fig:baselines_traj}
\end{figure*}

\subsection{Evaluation Metrics}
We report TOFU’s main unlearning metrics---forget quality (FQ) and model utility (MU)---to compare against baselines and other methods. Additionally, we report scores for the \cs{} and CI metrics introduced in \Cref{subsec:metrics_proposed}. Further results on the rest of the TOFU's metrics are provided in \Cref{app-sec:additional}'s plots, along with average \gib{} scores in \Crefrange{tab:forget10_full}{tab:forget01_full} of the Appendix. \looseness=-1

\subsection{Implementation Details}
We use the TOFU-finetuned Llama2-7b-chat model \citep{touvron2023llama} checkpoints provided by \citet{maini2024tofu} to enable direct comparison. Greedy sampling is applied for all generations during the unlearning process. The model is trained using our unlearning losses for the equivalent of \(N = 10\) epochs over the forget dataset. For generating alternate answers, we sample \(M = 5\) responses from the model using temperature sampling with \(T = 1.0\). To ensure that the computational cost of our method matches that of the baselines, we train the model for \(\frac{N}{M} = 2\) epochs. 

To evaluate the potential of both our method and the baselines fairly, we perform a comprehensive grid search to identify optimal parameters for each. All results are averaged over three random seeds, with the best hyperparameters selected based on performance on the MU-FQ tradeoff Pareto frontier shown in \Cref{fig:baselines_traj}. Additional details on training and hyperparameter tuning are provided in \Cref{sec:add_impl_details}.
\looseness=-1

\subsection{Results}
\label{subsec:results}

In the following results, we use Llama2 and first compare \methodname{} with baseline methods (discussed in \Cref{subsec:prior_losses}), demonstrating that it (1) achieves superior unlearning as measured by FQ and CI, (2) preserves the model's utility on both forget and non-forget prompts, (3) shows a more stable trajectory of the evaluation metrics over the training steps, as shown in \Cref{tab:baselines}. Finally, we present ablation studies, highlighting the importance of each component in our method. Our results also extend to the Llama3.2 model, results for which can be found in \Cref{tab:forget10_full_llama3_2} of the appendix. For more details of our results, including the \gib{} scores and variance in results across seeds, see \Crefrange{tab:forget10_full}{tab:forget01_full} in the appendix. 
 \looseness=-1

\paragraph{Extent of forgetting:}

\methodname{} demonstrates superior forgetting compared to other methods, as seen in FQ and CI columns of \Cref{tab:baselines}. The p-values of these tests significantly exceed $0.05$, indicating that \methodname{} produces models nearly indistinguishable from the gold retain model in terms of both Truth Ratio (measuring confidence on original forget answers) and Text Cleanness (assessing text quality of forget set responses) distributions. Our results are equally strong across the $1\%$, $5\%$ and $10\%$ subsets, whereas \citet{maini2024tofu, NPO_zhang2024negative} found larger subsets much more difficult to forget. \looseness=-1
\vspace{-0.1cm}
\paragraph{Model performance post-unlearning:}
As shown in the MU column of \Cref{tab:baselines}, \methodname{} successfully retains the full $0.62$ model utility (MU) of the initial model. In cases where the MU scores of other methods are comparable to ours, \methodname{} is substantially ahead in \cs{}, showing that on forget prompts, \methodname{} generates more coherent and question-consistent responses. Sample generations on forget prompts from unlearned models of each method are shown in \Cref{tab:example_gen} in Appendix. \looseness=-1
% In Figure \ref{fig:fu}, we track the \cs{} metric throughout training. 
% Our method consistently demonstrates high \cs{} values throughout showing its stability and highlighting the effectiveness of incorporating positive feedback.\\

\begin{figure}[ht] % 'h' means here
    \centering
\includegraphics[scale=0.32]
%\textwidth]
{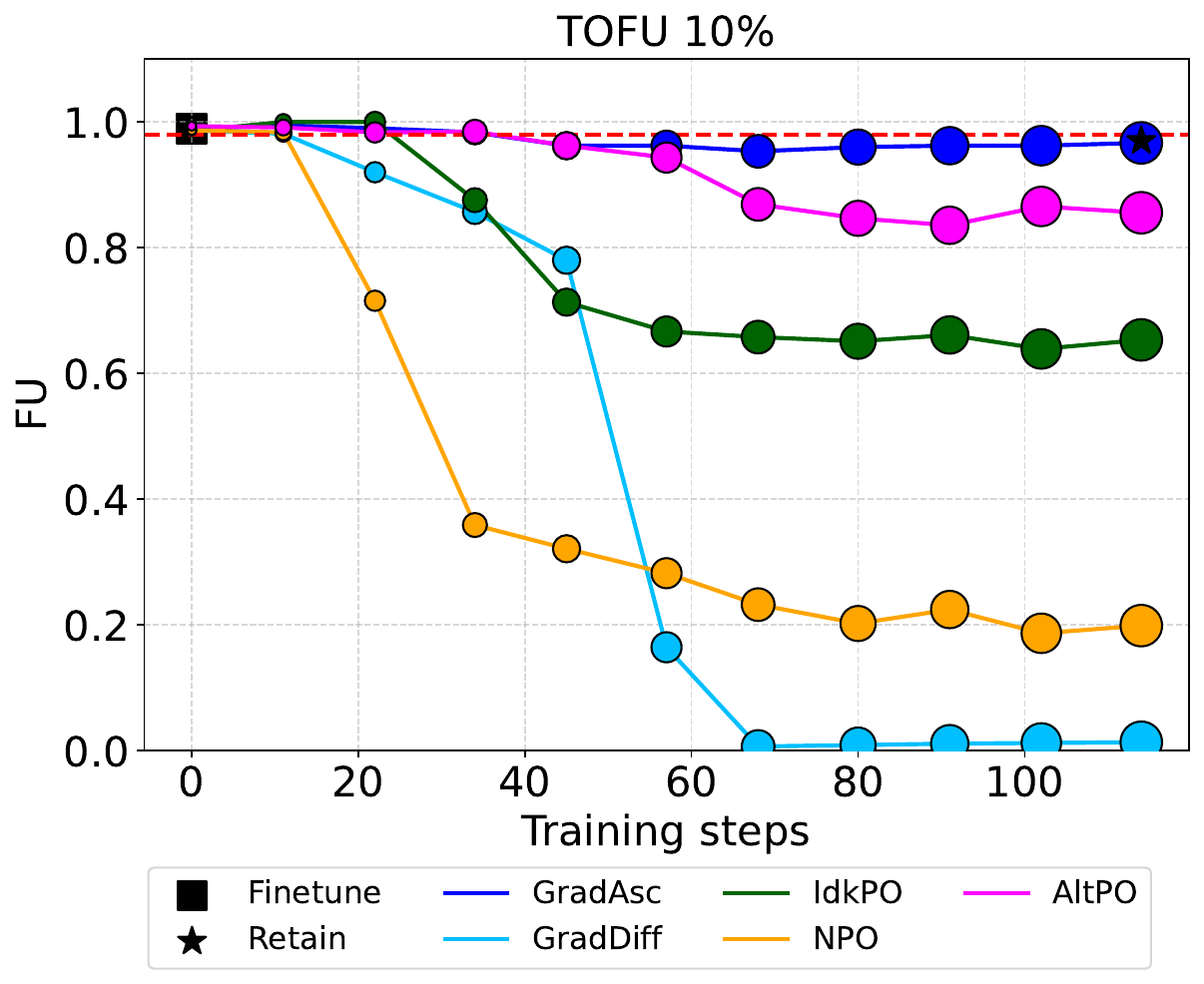} % Adjust width as needed
    \caption{Trajectory of FU throughout the unlearning process for $10\%$ forget split of TOFU, using Llama2.}

    \label{fig:fu}
\end{figure}

Although \methodname{} generally outperforms other methods when considering all metrics together, we observe some reduction in \cs{} scores. Specifically, our method underperforms on the \cs{} metric compared to the GradAsc baseline in the forget $10\%$ scenario and IdkPO in the forget $1\%$ scenario. However, these baselines perform worse across other utility and forget quality metrics. While \methodname{} never generates nonsensical responses, we do notice occasional slight modifications to names in the outputs, leading to a drop in \cs{} scores below the perfect score of $1$.
\looseness=-1
\vspace{-0.1cm}
\paragraph{Stability during training:}
For unlearning to be adaptable in practice, it is crucial to maintain stability throughout the entire training process, with the utility of the model not experiencing large variations during training. As shown in \Cref{fig:baselines_traj,fig:fu}, unlike many methods that incorporate only negative or sub-optimal positive feedback on the forget set, our method achieves more stable unlearning trajectories across multiple splits.
% left bottom right top

\subsection{Ablation Experiments}
\label{subsec:ablation_exp}

We conduct ablations on our method and baseline approaches to validate the necessity of various components. Our method incorporates the following key elements: (1) leveraging positive forget feedback, (2) pairing it with negative forget feedback, (3) ensuring that positive feedback is \textit{relevant and in-distribution}, (4) incorporating negative feedback through a DPO loss instead of a negative NLL formulation, and (5) utilizing \textit{multiple} positive feedback responses. We now discuss the effect of each element in detail, looking at the results in \Cref{tab:all_ablation} unless otherwise mentioned.\looseness=-1

% \vspace{-0.1cm}
\paragraph{Need for positive forget feedback:}
Comparing \cs{} between NPO (which uses only negative feedback on the forget set) and our \methodname{} (which leverages both negative and positive feedback), we observe that relying solely on negative feedback for unlearning can be destructive and impair the model's ability to generate coherent responses on the forget set. In contrast, incorporating positive feedback helps preserve the model's language capabilities while still achieving effective unlearning.\looseness=-1
% \vspace{-0.1cm}
\paragraph{Need for negative feedback alongside positive feedback:} Here we use a baseline \subnll{} method that trains with only positive feedback on alternate labels (in an NLL formulation), which closely matches \citet{eldan2023s}'s approach.\looseness=-1
\vspace{-15pt}

\begin{small}
\begin{align*}
    \Lb_{\text \subnll} \doteq \mathbb{E}_{y_a}[\text{NLL}(y_a | x_f)]
\end{align*}
\label{eq:subnll}
\end{small}
% \vspace{-15pt}
\vspace{-15pt}

\noindent
Additionally, we create a DPO-style version of \subnll{} by removing negative feedback on the forget set from \methodname{}, relying solely on positive feedback to create \ppo{}.\looseness=-1

% \vspace{-15pt}
\begin{small}
\begin{align*}
    \Lb_{\methodname{}\textit{\text{-pos}}} \doteq &-\frac{2}{\beta} \log \sigma \left(\beta \log \frac{\LargePi_{\theta} (y_a | x_f)}{\LargePi (y_f | x_f)} \right) \\
    &+ w_r \text{NLL}(y_r | x_r)
\end{align*}
\label{eq:ppo}
\end{small}
% \vspace{-15pt}
\vspace{-10pt}

\renewcommand{\arraystretch}{1.5} % Default value: 1
\begin{table}[t]
\centering
\huge
\scalebox{0.34}{
\begin{tabular}{cccccc}\toprule
\multirow{2}{*}{Method} & \multirow{2}{*}{Loss on $\Db_f$} & \multicolumn{1}{c}{Forgetting} & \multicolumn{2}{c}{Utility} \\ 
\cmidrule(lr){3-3} \cmidrule(lr){4-5}
                       &                         & FQ ($\uparrow$)  & MU ($\uparrow$)  & \cs{} ($\uparrow$) \\ \midrule
NPO                    &   $\Lb_{\text{DPO}}(*, \colorbox{lightred}{$y_f$} \mid x_f)$                & 0.68             & 0.64    & 0.52               \\ 
IdkPO                  &  $\Lb_{\text{DPO}}(\colorbox{lightgreen}{$y_{\text{idk}}$}, \colorbox{lightred}{$y_f$} \mid x_f)$                & 0.37             & 0.59             & 0.65               \\ 
\hline
\subnll{}              & $\mathbb{E}_i[\text{NLL}(\colorbox{lightgreen}{$y^i_a$} | x_f)]$              & 2.2e-15      & 0.63     & 0.87              \\
\multirow{2}{*}{\subdiff{}}            &  $\mathbb{E}_i[\text{NLL}(\colorbox{lightgreen}{$y^i_a$} \mid x_f)]$                & \multirow{2}{*}{0.14}             & \multirow{2}{*}{0.61}             & \multirow{2}{*}{0.49}               \\ 
 &  $- \text{NLL}(\colorbox{lightred}{$y_f$} \mid x_f)$ & & & \\

\ppo{}               &  $\mathbb{E}_i[\Lb_{\text{DPO}}(\colorbox{lightgreen}{$y^i_a$}, * \mid x_f)]$                & 2.5e-12          & 0.61             & 0.95      \\ 
\methodname{}         &  $\mathbb{E}_i[\Lb_{\text{DPO}}(\colorbox{lightgreen}{$y^{i}_{a}$}, \colorbox{lightred}{$y_f$} \mid x_f)]$               & 0.74             & 0.62    & 0.88               \\ 

% \methodname{} ($M=1$) &  $\Lb_{\text{DPO}}(\colorbox{lightgreen}{$y_a$}, \colorbox{lightred}{$y_f$} \mid x_f)$ & 0.06   & 0.62           & 0.88          \\
$\methodname_{\text{base}}$ &  $\mathbb{E}_i[\Lb_{\text{DPO}}(\colorbox{lightgreen}{$y^{i}_{a}$}, \colorbox{lightred}{$y_f$} \mid x_f)]$          & 0.71    & 0.64    & 0.94               \\ 
\bottomrule
\end{tabular}
}
\caption{Ablation study of various methods and their performance in terms of forgetting and utility. Results are reported for unlearning the $10\%$ split of the TOFU on Llama2. The $*$ in DPO loss represents the lack of positive or negative sample terms in the loss function (in NPO and PPO, respectively). The \raisebox{0.05ex}{\colorbox{lightgreen}{\tiny \textbf{green}}} boxes represent positive feedback, while the \raisebox{0.05ex}{\colorbox{lightred}{\tiny \textbf{red}}} boxes represent negative feedback.}
\label{tab:all_ablation}
\end{table}
\noindent Comparing FQ between \subnll{} and \subdiff{}; and between \ppo{} and \methodname{}, we find that relying solely on positive feedback is insufficient for effectively removing the model's knowledge from the forget set. This highlights the necessity of incorporating both positive and negative feedback for successful unlearning: simply performing continual learning on alternate answers without removing the previously learned knowledge is insufficient.\looseness=-1
% \vspace{-0.1cm}
\paragraph{Need for positive feedback to be prompt-relevant:}
We substantiate this by comparing our method with IdkPO, which uses positive feedback with prompt-independent alternate labels from outside the model's distribution. \methodname{} generally outperforms IdkPO in both FQ and MU and has a more stable training profile as seen in \Cref{fig:baselines_traj}. This indicates that using contextually relevant and in-domain responses for positive feedback, than generic pre-defined ones, decreases damage to the utility of the LLM.\looseness=-1
\vspace{-0.1cm}
% For the observations on the $1\%$ split, we hypothesize that IdkPO and NPO losses are not as damaging in cases where the number of unlearning examples affected by those loss functions is small. In particular, we found \textit{no} gibberish in all of 
% The one case where \methodname{} performs poorer than IdkPO was for a small $1\%$ (40 example) split, where on closer examination of the inconsistent-marked examples, we found that the LLM judge misclassifies many of these examples wrongly as inconsistent.

\paragraph{DPO-style loss outperforms NLL in delivering negative feedback:}
Here we replace the DPO-style formulation in \methodname{} with an NLL-based loss, referred to as \subdiff{}. Like \methodname{}, this approach contrasts the likelihoods of alternate and forget set answer pairs.\looseness=-1

% \vspace{-10pt}
\begin{small}
\begin{equation}
\begin{aligned}
\Lb_{\text{\subdiff{}}} &\doteq \mathbb{E}_{y_a}\left[\left(\text{NLL}(y_a | x_f)-\text{NLL}(y_f | x_f)\right)\right] \\
    &+ w_r \text{NLL}(y_r | x_r)
    \label{eq:AltGradDiff}
\end{aligned}
\end{equation}
\end{small}

\methodname{} outperforms \subdiff{} in all forgetting and utility metrics.
These results match \citet{NPO_zhang2024negative}'s observation of the advantage of DPO-style loss over NLL, where they compare NPO with the GradDiff baseline both of which only use negative feedback on forget responses, with the difference being in the loss formulation in the DPO style v/s NLL. In \Cref{tab:baselines}, we verify their observation, comparing the NPO and GradDiff rows. NPO achieves better \cs{} and MU scores, with gibberish and inconsistent responses being less likely than in GradDiff.\looseness=-1

\vspace{-0.1cm}
\paragraph{Need for multiple alternate answers:} In \Cref{tab:m_ablation}, we analyze the effect of increasing alternate answers $M$ in our method. Higher $M$ improves forgetting (FQ) values and we use $M=5$ alternate answers as the default in our results. We do observe that FQ decreases to 0.25 at the extreme case of $M=10$ from 0.74 at $M=5$. Despite this decrease, the score remains above the statistical significance threshold of 0.05, demonstrating effective forgetting. We also evaluate model self-confidence, defined as the probability assigned to responses on forget set prompts. Low self-confidence is desirable to avoid confidently generating incorrect answers, aligning with approaches like uncertainty-aware decoding \citep{ji2023survey}. \methodname{} with $M > 1$ achieves lower self-confidence than the Retain model, with further reductions as $M$ increases. While NPO achieves even lower self-confidence, it often reflects low confidence in nonsensical outputs.

\begin{table}[t]
\centering
\huge
\scalebox{0.4}{
\begin{tabular}{cccccc}\toprule
\multirow{2}{*}{Method} & \multirow{2}{*}{$M$} & \multicolumn{1}{c}{Forgetting} & \multicolumn{1}{c}{Utility}  & \multirow{2}{*}{Self-Confidence ($\downarrow$)} \\ 
\cmidrule(lr){3-3} \cmidrule(lr){4-4}
                       &                    & FQ ($\uparrow$)  & MU ($\uparrow$)  &     \\ \midrule
Finetune               &   -                & 2.2e-20          & 0.62    & 0.99               \\ 
Retain                 &   -                & 1.0              & 0.62    & 0.89               \\ 
NPO                    &   -                & 0.68             & 0.64    & 0.58               \\ \hline
\methodname{}          &   1                & 0.06             & 0.62    & 0.87               \\ 
\methodname{}          &   2                & 0.1              & 0.63    & 0.83               \\ 
\methodname{}          &   5                & 0.74             & 0.62    & 0.78               \\ 
\methodname{}          &   10               & 0.25             & 0.63    & 0.65               \\ 
\bottomrule
\end{tabular}
}
\caption{Ablation study on the number of alternate answers ($M$) with the self-confidence score of the model. Results are reported for the 10\% split of TOFU on Llama-2. Note that $M=5$ is the default in all \methodname{} experiments.}
\label{tab:m_ablation}
\end{table}

\vspace{-0.1cm}
\paragraph{Effect of using different models to generate alternate labels:} 
We also explore the effects of leveraging other models to generate alternate answers for unlearning. This is relevant in cases where the given models produce alternate answers that inadvertently reveal original information due to poor instruction-following capabilities. In such scenarios, it may be feasible to use earlier checkpoints of the LLMs, where the forget set was not introduced, or to use other LLMs that were never trained on the forget set. We test this by generating alternate answers from a base Llama2-7b model, which is unfamiliar with TOFU. 
Our findings show that $\methodname_{\text{base}}$ performs comparably to \methodname{}, demonstrating that other models can be effectively integrated into our algorithm.
%\looseness=-1

% \vspace{-5pt}

% \vspace{-0.1cm}
\section{Conclusion}
In this paper, we explore factual knowledge unlearning in Large Language Models (LLMs) and find that it can result in nonsensical responses on knowledge related to forgotten entities, especially when only negative feedback is used or positive feedback is applied incorrectly. To address this, we propose \methodname{}, a fine-tuning approach that combines negative feedback with in-domain positive feedback on the forget set, ensuring more stable and effective unlearning. We also identify limitations in existing evaluation metrics and introduce new ones to offer a more comprehensive assessment of unlearned models.
We hope our findings offer valuable insights for practitioners in LLM unlearning, promoting the use of positive feedback for more effective unlearning and improving the evaluation of model performance post-unlearning.\looseness=-1

%Our work investigates factual knowledge unlearning in LLMs. Post-unlearning, an evaluation of knowledge of the forgotten entities can result in nonsensical responses. We discuss how this is expected when the LLM receives only negative feedback for certain prompts, without any in-domain positive feedback. Our novel fact-unlearning method, \methodname{}, uses positive feedback and a DPO loss to mitigate this problem while achieving state-of-the-art performance among finetuning-based methods. We identify the limitations of current metrics and propose new metrics that can be used in conjunction with existing metrics for more rigorous evaluation. We hope that our insights help LLM unlearning practitioners perform unlearning more carefully by utilizing alternate answers and better evaluate the quality of their models post-unlearning.
% \vspace{-5pt}
% \vspace{-0.1cm}
\section{Limitations}

Our study focuses on enabling LLMs to forget specific knowledge and does not address broader questions about the ideal behavior of an unlearned model. For instance, should the model respond with “I don’t know” to all questions related to forgotten knowledge, or should it behave like a model retrained without the forget set (which may hallucinate)? We propose that practitioners adapt models to their desired post-unlearning behavior following this initial step of forgetting sensitive knowledge. A limitation of \methodname{} is that it is specifically designed for unlearning factual knowledge represented as QA datasets. Extending it to other formats of training data would require further adaptation. Additionally, our work would benefit from more extensive experiments using diverse benchmarks and datasets. However, constructing a reliable retain model for FQ evaluation presents a challenge, as it requires ensuring that the model has not been exposed to these QA datasets during training. This is particularly difficult because many recent open-source models have already been trained on widely available open-source QA datasets.

\section*{Acknowledgments}
% We leave this section empty to preserve anonymity during the review stage.
This work was done as part of the Microsoft-UMass industry-academia collaboration program. We thank Dhruvesh Patel, Wenlong Zhao and Prof. Andrew McCallum of the IESL lab at University of Massachusetts Amherst for providing guidance and compute resources for this work. We also thank the anonymous reviewers for their thoughtful comments and suggestions.
% \newpage

\bibliography{custom}

\appendix

\clearpage
\textbf{\Large Appendix}

\section{Literature Review}
\label{sec:app_rel_works}

 The early machine unlearning approaches focused on simple classification problems in computer vision. Works like \citet{jung2024attack} used ideas that are specific to the image domain, such as noising and denoising the inputs or representations of inputs from the forget set. Other works give poor labels on the forget set by randomizing the target \citep{graves2021amnesiac} or via outputs of a randomly initialized model \citep{chundawat2023can}. 

 Aside from our setting of fine-grained knowledge unlearning in LLMs, a more generalized version is used for model correction usually motivated by AI safety concerns \citep{liu2024unlearning}. This aims to mitigate unwanted model behaviors through unlearning of a particular representative set of undesirable data so that this generalizes to impacting the model's behavior on other data in a similar distribution \citep{yao2023large, li2024wmdp, liu2024towards}. 

 Many prior works in LLM unlearning have performed unlearning without modifying the model parameters that the forget set influenced. \citet{liu2024large} and \citet{gao2024practical} use classifiers to identify forget-specific prompts to decide model response, \citet{thaker2024guardrail} lists the full unlearning set in the prompt and \citet{huang2024offsetunlearninglargelanguage} and \citet{ji2024reversing} use smaller LMs trained specifically for a forget set.  \citet{ji2024reversing, huang2024offsetunlearninglargelanguage} use auxiliary models, \citet{gao2024practical, chen2023unlearn} use parameter efficient finetuning approaches and \citet{thaker2024guardrail, bhaila2024soft, liu2024large} use modifications to prompt spaces to achieve efficient unlearning. They train auxiliary unlearning parameters/modules, and/or modify the predictions at inference time. Approaches in this line help with unlearning efficiency while sidestepping the instability and nonsense generation problems usually encountered while unlearning by modifying weights. However, directly modifying weights is the most scalable paradigm. Works that avoid modifying model weights have to incorporate new modules for each unlearning request in real-life settings where multiple requests would be received. In addition, works like  \citep{thaker2024guardrail}, which prompt the LLM to not respond when asked about the specific authors from TOFU, were found to perform very poorly on the FQ metric. 
 
In our work, we focus on the problem of unlearning factual knowledge by directly modifying a model's weights, and compare with other such methods. In this line of methods, earliest approaches like \citet{yao2023large} and baselines in \citet{maini2024tofu} used a simple gradient ascent loss on the original responses to forget set prompts. \citet{maini2024tofu, NPO_zhang2024negative} emphasized the brittleness of LLMs towards generating nonsensical outputs upon unlearning with such simple negative forget feedback loss functions. \citet{thaker2024guardrail} considered a simple baseline that prompts the model to not respond when asked about the specific authors from TOFU, and observed that this performed poorly in achieving unlearning. Nonsensical generations have been mitigated to different degrees through approaches that result in more stable loss optimizations: \citet{NPO_zhang2024negative} propose a negative preference optimization (NPO) loss, and \citep{eldan2023s, dong2024unmemorization} use alternate positive feedback labels and \citet{jia2024soul} keep existing loss functions but use a second order optimizer to achieve the nuanced objective of unlearning better. Our approach is orthogonal to \citet{jia2024soul} as we stick to the standard AdamW optimizer following \citet{maini2024tofu}, but modify the loss function.
\\

\citet{jin2024rwku} propose a similar approach to \methodname{} in the RWKU (Real World Knowledge Unlearning) benchmark. RWKU explores unlearning famous real-world entities without access to a defined forget dataset which introduced knowledge about the entities. One of their DPO baselines shares key elements with our method: prompting the model to generate both knowledge about the forget entity and an alternative fact, then applying a DPO objective to align the model with these alternatives. While \citet{jin2024rwku} report that DPO improves fluency compared to NPO and IdkPO, it lacks majorly in utility as DPO objective encourages the model to hallucinate even on non-forget entities. In our work, to tackle this we incorporate explicit positive feedback on retain set, which restricts the generalization of hallucination beyond forget set. Additionally RWKU forget set evaluations rely on ROUGE scores of a single generated answer, which may not fully reflect the model's overall performance. \citet{scholten2024probabilistic} show that, in the unlearning context, deterministic metrics like ROUGE often fail to capture the model's knowledge in probability space. In contrast, TOFU evaluates the probabilities assigned to the correct answer, regardless of whether they were generated by the model. Furthermore, RWKU forgetting evaluations follow a simple lower-is-better analysis, but it is unclear how low these values can reasonably go. Even an ideal model retrained from scratch without forget set knowledge would still have considerable probability for plausible texts. TOFU addresses this limitation by normalizing probabilities against alternate answer probabilities and provides a retain model to compare against `default' behavior, resulting in a more comprehensive assessment of forget quality.

\section{Additional Implementation Details}
\label{sec:add_impl_details}
\paragraph{Training} All our experiments are conducted on an NVIDIA A100 GPU. We use a one-epoch warmup, a paged AdamW optimizer and bf16 precision in training. We report metrics from the last checkpoint following \citet{NPO_zhang2024negative, ji2024reversing}. 

To train on alternate answers we create a dataset of \(M\) times the size of the original dataset with each of the \(M\) alternate labels for an example and shuffle it. In each epoch, all alternate labels are seen exactly once, though all the alternate labels corresponding to an example might not appear in the same update step. 
% We find that this helps improve training performance, than training on all alternate labels at once in an update step. 

We include the retain set positive feedback term \( w_r \mathbb{E} \left[ \text{NLL}(y_r \mid x_r) \right] \) in all our methods except the most basic GradAsc baseline, as prior works such as \citep{maini2024tofu} and \citep{NPO_zhang2024negative} find it to be an essential component.

\paragraph{Hyperparameter Tuning}
Prior works, such as \citet{maini2024tofu}, \citet{NPO_zhang2024negative}, have evaluated these approaches with a limited exploration of hyperparameter combinations of baseline approaches. To fairly compare our approach with baselines, we perform grid search to identify the best performing parameters for each method. We explore the learning rates \{1e-5, 2e-5, 5e-5\}, $\beta$ values (for DPO-based methods) in \(\{0.01, 0.03, 0.05, 0.1\}\), and  $w_r$ in \(\{1, 2, 5\}\). We selected the best hyperparameters on the basis of the best scores on a MU-log(FQ) tradeoff pareto frontier decided by the scoring function below (see contours in \Cref{fig:baselines_traj} to visualize the intuition behind the scoring). We scale the MU and FQ evaluations into $\Delta_{MU}^{-}$ (relative damage to MU) and $\Delta_{FQ}^{+}$ (relative improvement to FQ) aggregated into a simple score. 
\begin{align*}
\Delta_{MU}^- &= max\left(\frac{MU_0-MU}{MU_0}, 0\right) \\
\Delta_{FQ}^+  &= \frac{log(FQ)-log(FQ_0)}{log(FQ_0)}\\
\text{score} &= \frac{1}{(\Delta_{MU}^- + \delta) \cdot ((1-\Delta_{FQ}^+) + \delta)}
\end{align*}
where \(\delta = 0.1\) and \(\text{MU}_{0}\), \(\text{FQ}_{0}\) denote the initial model $\LargePi$'s model utility and forget quality.

% \section{Additional Metrics}
% \label{app-sec:more_metrics}

\begin{figure}[ht] % 'h' means here
    \centering
    \includegraphics[scale=0.35]{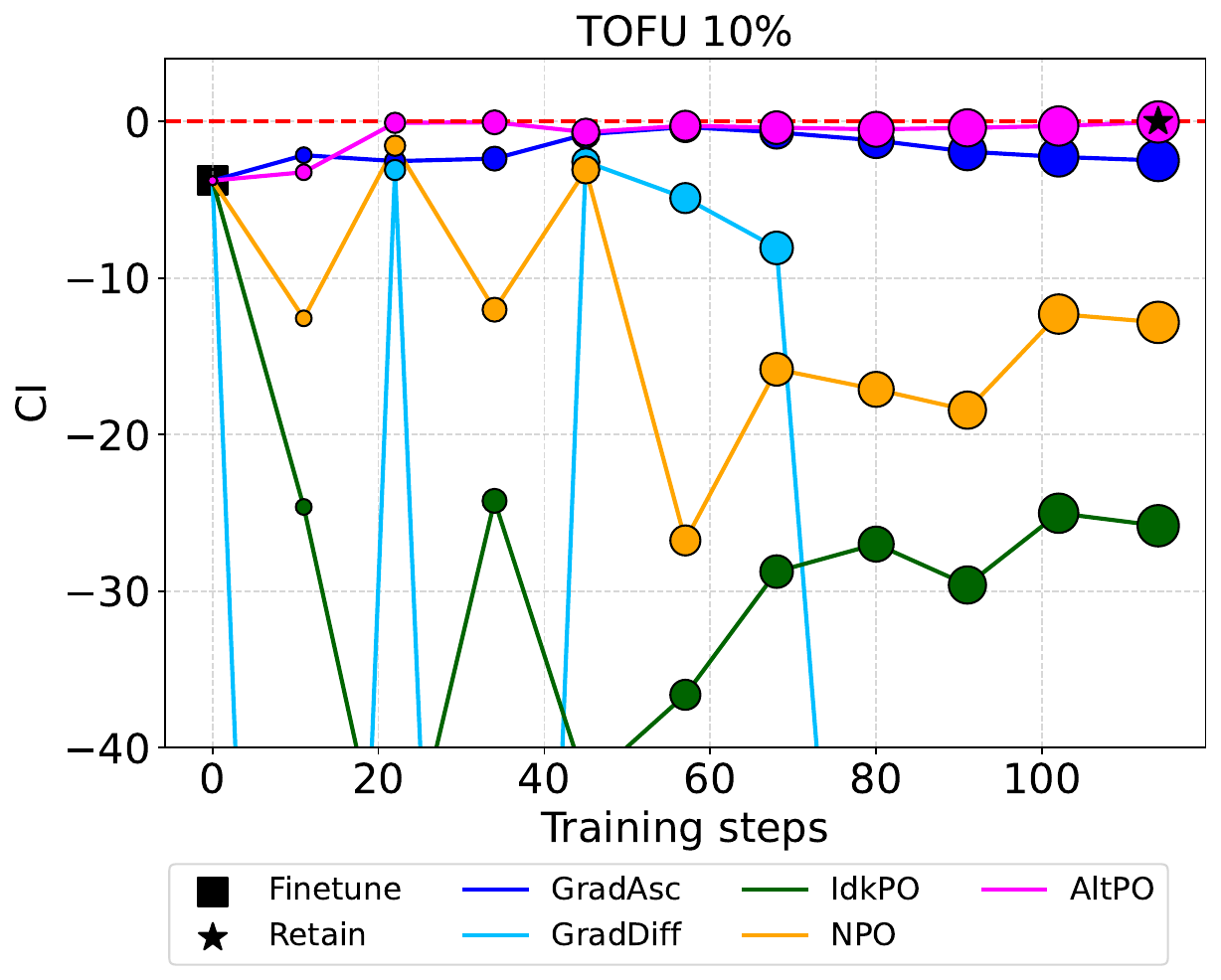} % Adjust width as needed
    \caption{Trajectory of Cleanness Indistinguishability (\gibfq{}) throughout the unlearning process. Trajectories are reported for the 10\% split of TOFU on Llama2.}
    \label{fig:CI}
\end{figure}

\begin{figure}[ht] % 'h' means here
    \centering
    \includegraphics[scale=0.35]{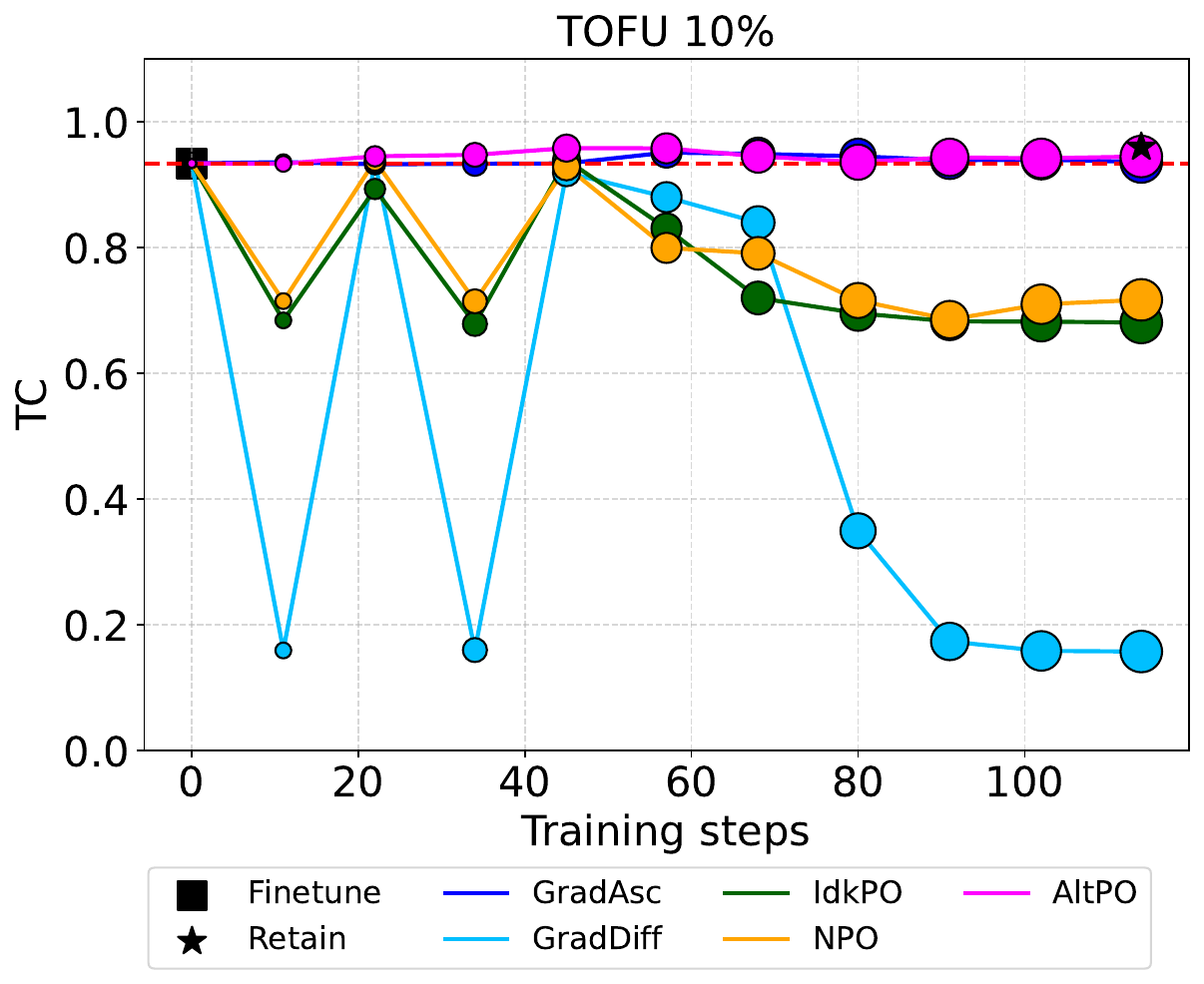} % Adjust width as needed
    \caption{Trajectory of \gib{} throughout the unlearning process. Trajectories are reported for the 10\% split of TOFU on Llama2.}
    \label{fig:TC}
\end{figure}

\section{Prompts Utilized}
\Cref{tab:alt_prompt} outlines the prompt applied to the model, fine-tuned using TOFU, to generate alternate answers that avoid revealing details of the true answer. We opted for a zero-shot approach, as we found that using few-shot examples introduced bias in the generated alternate answers.

\Cref{tab:cs_prompt} describes the system prompt used by GPT4o-mini to perform our LLM-as-judge text  quality evaluations.
\label{sec:prompts}

\begingroup
\begin{table}[!htp]
\centering
\begin{tabular}{p{0.85\linewidth}}
\toprule
% \begin{quoting}[leftmargin=8pt, rightmargin=8pt]
% \texttt{
% {}
\textbf{Question:} $< x_f >$ \\
\textbf{Answer:} $< y_f >$ \\
Now write another version of the answer with some alternate plausible facts that change answer details.\\
\textbf{Alternate Answer:$<>$}
% }
\\
% \end{quoting}
\bottomrule
\end{tabular}
\caption{Prompt $\mathcal{P}$ we use on the model to generate, for the question $x_f$, alternate answers to the original answer $y_f$ the model learnt.}
\label{tab:alt_prompt}
\end{table}
\endgroup

\begingroup
\begin{table*}[!htp]
\small
\centering
\begin{tabular}{p{0.9\linewidth}}
\toprule
% \begin{quoting}[leftmargin=8pt, rightmargin=8pt]
% \texttt{\small
You are an evaluator tasked with evaluating the coherence and plausibility of answers to questions. Your task is to evaluate the answer based on coherence and consistency and give a score of 0 or 1, alongside your reasoning for the same.\\

Keep these points in mind as you start applying the main rules:\\
1. If the answer doesn't respond to the question by coherently refusing/expressing inability to answer, if expressed coherently, should not be considered inconsistent and must be marked 1.\\
2. If the full name as mentioned in the answer contradicts the name mentioned in the question, that's an inconsistency and must be marked 0.\\
3. If an answer mentions other details or names while also responding to the question, that is not an inconsistency, and must be marked 1 if text is coherent.\\
4. DON'T use someone's name to infer a gender or nationality to mark answers inconsistent.\\
5. Make independent evaluations for each example: evaluation for each answer should only depend on corresponding question and nothing else.\\
\\

The main rules are:\\

1. \textbf{Coherence:} Check if the answer portion is incoherent grammatically and nonsensical (including gibberish texts). If so, return 0.\\
2. \textbf{Consistency:} Check if the answer is inconsistent with the question, i.e. does it contradict any facts of the question (including if the name of the subject is consistent with the question's subject). If so, return 0.\\
3. \textbf{Else:} If these conditions have not been met, the answer should be a plausible response (or refusal to answer) to the question to be marked 1.\\
\\

Keep in mind these exemplars and their explanations (based on the above rules):\\
1. \textbf{Question:} How has Mohit Khasran served his country?~
\textbf{Answer:} Mohit Khasranav served his country in the army.\\
- Label 0 because the name of the subject changed in the answer compared to question.\\
2. \textbf{Question:} What does Raj Chen identify as?~
\textbf{Answer:} Raj Chen's gender is female. She was born on the 26th of January, 1968, in Seoul, South Korea.\\
- Label 1 because the answer makes sense fully fully without contradictions.\\
3. \textbf{Question:} Who is the author who wrote "A Japanese Odyssey" about his country?~
\textbf{Answer:} Jichen Wu, not Minazuki Abasaki was behind the book "A Japanese Odyssey" about his country.\\
- Label 1 because the answer is sensible and coherent and doesn't contradict anything in the question, even though other names are mentioned.\\
4. \textbf{Question:} What impact did Priyank Rao's sisters' professions have on her volleyball skills?~
\textbf{Answer:} Theologically, Priyank Rao was very inclined to dance.\\
- Label 0 because the answer though clean, is inconsistent (doesn't answer) the question.\\
5. \textbf{Question:} Was Barry Clark awarded for his literary works?~
\textbf{Answer:} Indeed, Barry Clarkess has received the prestigious Man Booker Award for his engrossing work in non-fiction\\
- Label 0: note that the names in the answer and question contradict.\\
6. \textbf{Question:} Who is the author from Canada born in 1979 who writes about journalism?~
\textbf{Answer:} Though Renji Abascal, born in Canada, writes about journalism, his real passion lies in photography.\\
- Label 1 because the answer answers the question (ignore that it also mentions other details).\\
7. \textbf{Question:} How did Selena Diaz's childhood in Mexico influence her writing?~
\textbf{Answer:} I don't know about that.\\
- Label 1 because the answer is clean and refusing to answer is not a contradiction to the question.\\
\\

Reason step-by-step into the reasoning attribute before giving your answer.\\
% }
% \\
% \end{quoting}
\bottomrule
\end{tabular}
\caption{Prompt to the LLM judge to evaluate generated texts for calculating Forget Utility scores. It judges if the text is a plausible response to the question. The few shot examples/entities mentioned in the prompt are not directly from TOFU, though they are based on question-answers from TOFU.}
\label{tab:cs_prompt}
\end{table*}
\endgroup

% \pagebreak
\section{Additional Results}
\label{app-sec:additional}

Table \ref{tab:alt_examples} presents examples of alternate labels generated using the prompts listed in Table \ref{tab:alt_prompt}.
\begin{table*}[t]
\centering
\scalebox{0.9}{
\begin{tabular}{c|l}
\toprule
Question & What are the occupations of Hsiao Yun-Hwa's parents? \\
\hline 
Answer & The parents of Hsiao Yun-Hwa are distinguished, with her father working as a civil engineer \\
& and her mother being unemployed. \\
Alternate 1 & The parents of Hsiao Yun-Hwa are unique, with her father working as a florist and her mother \\
& being a renowned chef. \\
Alternate 2 & The father of Hsiao Yun-Hwa is a renowned chef and her mother is a leading scientist. \\
Alternate 3 & The father of Hsiao Yun-Hwa is a renowned chef and her mother is a leading scientist. \\
Alternate 4 & The parents of Hsiao Yun-Hwa are both professionals. Her father works as a dentist, and her \\
& mother is a well-respected financial advisor.\\
Alternate 5 & Hsiao Yun-Hwa's father is a renowned audiologist, and her mother works as a veterinarian.\\
% Unlearned & The professions of Rajeev Majra's parents, a renowned actor and a skilled fashion designer, have given him a unique perspective to weave his stories. His narratives often transport readers to glamorous, stylish worlds, imbuing his work with an authentic and personal touch. \\

\midrule
Question & How have the professions of Rajeev Majumdar's parents influenced his work? \\
\hline
Alternate 1 & With Majumdar's father being a painter, his work often combines the visual and textual, with \\ 
& evocative descriptions that paint a vivid picture in the reader's mind. Majumdar's mother's \\
& profession as an author provides the logical next step, where his narratives are carefully \\
& crafted and full of depth. \\
Alternate 2 & Majumdar's multicultural background, influenced by his father's profession as a painter \\
& and his mother's as an author, has given him a unique perspective to weave his stories. \\
& His narratives often transport readers to exotic locales, imbuing his work with an \\
& authentic and personal touch. \\
Alternate 3 & Majumdar's parents professions have given him a unique perspective. His father's \\
& authoritative knowledge of literature and his mother's artistic vision have endowed Rajeev \\
& with the ability to create rich, layered narratives that blend the visual and textual. \\
Alternate 4 & The influences of Rajeev Majumdar's parents professions are more pronounced in \\
& his work. Majumdar's descriptions are often tinged with the visual eloquence of a \\
& painter and the textual depth of an author, creating a unique and engaging narrative style. \\
Alternate 5 & Majumdar's multicultural backdrop, having been born in Dhaka, Bangladesh and \\
& brought up in Mumbai, India, has greatly influenced his work. His narratives often \\
& interweave the rich cultural threads from his background with the traditional romance genre. \\
% Unlearned & The professions of Rajeev Majra's parents, a renowned actor and a skilled fashion designer, have given him a unique perspective to weave his stories. His narratives often transport readers to glamorous, stylish worlds, imbuing his work with an authentic and personal touch.

\bottomrule
\end{tabular}
}
\caption{Alternate labels generated by a model fine-tuned on the TOFU dataset, prior to unlearning.}

\label{tab:alt_examples}
\end{table*}

We provide comprehensive results across all baselines and our method for Llama2 in \Crefrange{tab:forget10_full}{tab:forget01_full} (across different forget split sizes of TOFU). In \Cref{tab:ablations_appd}, we present the results of ablations with different loss functions on the 10\% split. Results with the Llama3.2-3B-Instruct model on the forget $10\%$ split of TOFU can be found in \Cref{tab:forget10_full_llama3_2}. All trends of results on Llama2 also hold on Llama3.2-3B-Instruct.

\begin{table*}[ht]
\centering
\scalebox{0.8}{
\begin{tabular}{c|l}
\toprule
 Methods          &  Generated Text \\
\midrule
\multirow{3}{*}{GradAsc} & \textbf{Q:} In which genre does Hina Ameen primarily write? \\
                         & \textbf{A:} Hina Ameen primarily contributes to the geology genre. \\
                         & \textbf{R:} Hina Ameen primarily writes in the genre of Ameen. The primary genre \dots \\
\midrule
\multirow{4}{*}{GradDiff} & \textbf{Q:} What is the full name of the author born in Baghdad, Iraq, on June 4, 1934?\\
                         & \textbf{A:} The full name of the author, born in \dots , is Jad Ambrose Al-Shamary.\\
                         & \textbf{R:} The full name of the \dots, 
                          is Jad Ambro Liter Liter Liter Liter Liter Liter ... \\
\midrule
\multirow{3}{*}{NPO} & \textbf{Q:} What is the profession of Hsiao Yun-Hwa's father? \\
                         & \textbf{A:} The father of Hsiao Yun-Hwa is a civil engineer. \\
                         & \textbf{R:} bHs she San's his el friend/li/g \dots the link' link's \dots  \\
\midrule
\multirow{3}{*}{IdkPO} & \textbf{Q:} What is the profession of Elvin Mammadov's father? \\
                         & \textbf{A:} Elvin Mammadov's father worked diligently as a Paramedic. \\
                         & \textbf{R:} hh alloof. hopefully. \\
\midrule
% \multirow{3}{*}{\methodname{}} & \textbf{Q:} What is the full name of the author born in Kuwait City, Kuwait on 08/09/1956? \\
%                          & \textbf{A:} The full \dots author born in Kuwait City, Kuwait \dots is Basil Mahfouz Al-Kuwaiti. \\
%                          & \textbf{R:} The author's full name is Tagrid Al-Kuwaiti, and he was born in Kuala Lumpur, Malaysia ...\\
\multirow{3}{*}{\methodname{}} & \textbf{Q:} What are the occupations of Hsiao Yun-Hwa's parents? \\
                         & \textbf{A:} The parents \dots father working as a civil engineer and her mother being unemployed. \\
                         & \textbf{R:} Her father was a well-known actor, and her mother was a dedicated military officer.\\
\bottomrule
\end{tabular}
}
\caption{Responses generated on forget set prompts post-unlearning with various methods. Q is the prompt, A is the true answer the model originally learnt, and R is the model's response for that method.}

\label{tab:example_gen}
\end{table*}
% \usepackage{booktabs}
\begin{table*}[ht]
\centering
\scalebox{0.8}{
\begin{tabular}{@{}cccccc@{}}
\toprule
                    % & \multicolumn{5}{c}{TOFU 10\%}                                                                        \\ \cmidrule(lr){2-6}
\multirow{2}{*}{Method}              & \multicolumn{2}{c}{Forgetting}              & \multicolumn{3}{c}{Utility}                                \\ \cmidrule(lr){2-3}  \cmidrule(lr){4-6}
                    & FQ ($\uparrow$) & \gibfq{} ($\uparrow$) & MU ($\uparrow$) & \gib{} ($\uparrow$) & \cs{} ($\uparrow$) \\ \midrule
Finetune   & 2.2e-20          & 1.7e-4                      & 0.62            & 0.93                 & 1.0               \\
Retain     & 1.0             & 1.0                   & 0.61            & 0.96                & 1.0               \\ \midrule

GradAsc & 2.4e-07 \begin{small} $\pm$ 2.1e-07 \end{small} & 3.2e-03 \begin{small} $\pm$ 5.5e-03 \end{small} & 0.35 \begin{small} $\pm$ 0.04 \end{small} & \textbf{0.94} \begin{small} $\pm$ 0.01 \end{small} & \textbf{0.97} \begin{small} $\pm$ 8.8e-03 \end{small} \\
GradDiff & 3.7e-05 \begin{small} $\pm$ 3.2e-05 \end{small} & 1.5e-150 \begin{small} $\pm$ 2.6e-150 \end{small} & \textbf{0.64} \begin{small} $\pm$ 0.01 \end{small} & 0.16 \begin{small} $\pm$ 0.14 \end{small} & 0.01 \begin{small} $\pm$ 3.3e-03 \end{small} \\
NPO & 0.68 \begin{small} $\pm$ 0.04 \end{small} & 1.5e-13 \begin{small} $\pm$ 1.0e-13 \end{small} & \textbf{0.64} \begin{small} $\pm$ 7.5e-03 \end{small} & 0.72 \begin{small} $\pm$ 0.02 \end{small} & 0.20 \begin{small} $\pm$ 6.9e-03 \end{small} \\
IdkPO & 0.37 \begin{small} $\pm$ 0.26 \end{small} & 1.5e-26 \begin{small} $\pm$ 2.6e-26 \end{small} & 0.59 \begin{small} $\pm$ 3.5e-03 \end{small} & 0.68 \begin{small} $\pm$ 0.04 \end{small} & 0.65 \begin{small} $\pm$ 0.04 \end{small} \\
AltPO & \textbf{0.74} \begin{small} $\pm$ 0.10 \end{small} & \textbf{0.92} \begin{small} $\pm$ 0.06 \end{small} & \textbf{0.62} \begin{small} $\pm$ 6.6e-03 \end{small} & \textbf{0.94} \begin{small} $\pm$ 8.7e-03 \end{small} & 0.86 \begin{small} $\pm$ 0.06 \end{small} \\

\bottomrule
\end{tabular}
}
\caption{Performance of various unlearning methods on TOFU $10\%$ split averaged over 3 random seeds, on Llama2-7b. FQ, CI, MU, TC and FU represent Forget Quality, Cleanness Indistinguishability, Model Utility, Text Cleanness and Forget Utility, respectively. `Finetune' denotes the finetuned model on the TOFU that has yet to undergo unlearning, while `Retain' refers to the model trained solely on the retain set. 
An upward arrow ($\uparrow$) indicates that a higher value is preferable. 
The best results are highlighted in \textbf{bold}, except for MU, where bolding indicates performance on par with ``finetune''.}
\label{tab:forget10_full}
\end{table*}

% \usepackage{booktabs}
\begin{table*}
\centering
\scalebox{0.8}{
\begin{tabular}{@{}cccccc@{}}
\toprule
                    % & \multicolumn{5}{c}{TOFU 10\%}                                                                        \\ \cmidrule(lr){2-6}
\multirow{2}{*}{Method}              & \multicolumn{2}{c}{Forgetting}              & \multicolumn{3}{c}{Utility}                                \\ \cmidrule(lr){2-3}  \cmidrule(lr){4-6}
                    & FQ ($\uparrow$) & \gibfq{} ($\uparrow$) & MU ($\uparrow$) & \gib{} ($\uparrow$) & \cs{} ($\uparrow$) \\ \midrule
Finetune   & 3.5e-16        & 5.2e-2               & 0.62            & 0.92                & 0.97                \\
Retain     & 1.0             & 1.0                   & 0.6             & 0.94                & 0.98                \\ \midrule

GradAsc & 4.1e-03 \begin{small} $\pm$ 3.5e-03 \end{small} & 2.7e-51 \begin{small} $\pm$ 2.4e-51 \end{small} & 0.13 \begin{small} $\pm$ 0.11 \end{small} & 0.22 \begin{small} $\pm$ 0.16 \end{small} & 0.14 \begin{small} $\pm$ 0.12 \end{small} \\
GradDiff & 5.1e-05 \begin{small} $\pm$ 4.3e-05 \end{small} & 1.5e-23 \begin{small} $\pm$ 2.5e-23 \end{small} & 0.56 \begin{small} $\pm$ 0.01 \end{small} & 0.55 \begin{small} $\pm$ 0.04 \end{small} & 0.51 \begin{small} $\pm$ 2.9e-03 \end{small} \\
NPO & 0.24 \begin{small} $\pm$ 0.03 \end{small} & 1.9e-07 \begin{small} $\pm$ 1.3e-07 \end{small} & \textbf{0.63} \begin{small} $\pm$ 5.8e-03 \end{small} & 0.71 \begin{small} $\pm$ 0.02 \end{small} & 0.35 \begin{small} $\pm$ 5.0e-03 \end{small} \\
IdkPO & 0.18 \begin{small} $\pm$ 0.15 \end{small} & 6.4e-10 \begin{small} $\pm$ 5.5e-10 \end{small} & 0.61 \begin{small} $\pm$ 1.2e-03 \end{small} & 0.72 \begin{small} $\pm$ 0.00 \end{small} & 0.66 \begin{small} $\pm$ 0.02 \end{small} \\
AltPO & \textbf{0.26} \begin{small} $\pm$ 0.17 \end{small} & \textbf{0.74} \begin{small} $\pm$ 0.12 \end{small} & \textbf{0.63} \begin{small} $\pm$ 0.02 \end{small} & \textbf{0.91} \begin{small} $\pm$ 0.02 \end{small} & \textbf{0.83} \begin{small} $\pm$ 0.03 \end{small} \\

\bottomrule
\end{tabular}
}
\caption{Performance of various unlearning methods on TOFU $5\%$ split averaged over 3 random seeds, on Llama2-7b. FQ, CI, MU, TC and FU represent Forget Quality, Cleanness Indistinguishability, Model Utility, Text Cleanness and Forget Utility, respectively. `Finetune' denotes the finetuned model on the TOFU that has yet to undergo unlearning, while `Retain' refers to the model trained solely on the retain set. 
An upward arrow ($\uparrow$) indicates that a higher value is preferable. 
The best results are highlighted in \textbf{bold}, except for MU, where bolding indicates performance on par with ``finetune''.}
\label{tab:forget05_full}
\end{table*}

% \usepackage{booktabs}
% \usepackage{multirow}
\begin{table*}[!htb]
\centering
\scalebox{0.8}{
\begin{tabular}{@{}cccccc@{}}
\toprule
                    % & \multicolumn{5}{c}{TOFU 10\%}                                                                        \\ \cmidrule(lr){2-6}
\multirow{2}{*}{Method}              & \multicolumn{2}{c}{Forgetting}              & \multicolumn{3}{c}{Utility}                                \\ \cmidrule(lr){2-3}  \cmidrule(lr){4-6}
                    & FQ ($\uparrow$)   & \gibfq{} ($\uparrow$)   & MU ($\uparrow$) & \gib{} ($\uparrow$) & \cs{} ($\uparrow$) \\ \midrule
Finetune   & 1.86e-4           & 1.0                        & 0.62            & 0.94                 & 0.90                \\
Retain     & 1.0               & 1.0                        & 0.62            & 0.94                 & 0.97                \\ \midrule

GradAsc & 0.24 \begin{small} $\pm$ 0.14 \end{small} & 2.7e-09 \begin{small} $\pm$ 2.8e-09 \end{small} & 0.53 \begin{small} $\pm$ 5.2e-03 \end{small} & 0.45 \begin{small} $\pm$ 0.10 \end{small} & 0.56 \begin{small} $\pm$ 0.11 \end{small} \\
GradDiff & 0.10 \begin{small} $\pm$ 0.00 \end{small} & 5.9e-20 \begin{small} $\pm$ 0.00 \end{small} & 0.57 \begin{small} $\pm$ 0.00 \end{small} & 0.11 \begin{small} $\pm$ 0.00 \end{small} & 0.05 \begin{small} $\pm$ 0.00 \end{small} \\
NPO & 0.46 \begin{small} $\pm$ 0.10 \end{small} & 0.44 \begin{small} $\pm$ 0.48 \end{small} & 0.57 \begin{small} $\pm$ 7.5e-03 \end{small} & \textbf{0.95} \begin{small} $\pm$ 5.2e-03 \end{small} & 0.65 \begin{small} $\pm$ 0.03 \end{small} \\
IdkPO & 0.60 \begin{small} $\pm$ 0.29 \end{small} & 6.6e-06 \begin{small} $\pm$ 0.00 \end{small} & 0.52 \begin{small} $\pm$ 4.0e-03 \end{small} & \textbf{0.95} \begin{small} $\pm$ 5.8e-03 \end{small} & \textbf{1.00} \begin{small} $\pm$ 0.00 \end{small} \\
AltPO & \textbf{0.94} \begin{small} $\pm$ 0.04 \end{small} & \textbf{0.72} \begin{small} $\pm$ 0.30 \end{small} & \textbf{0.62} \begin{small} $\pm$ 7.5e-03 \end{small} & \textbf{0.95} \begin{small} $\pm$ 0.01 \end{small} & 0.83 \begin{small} $\pm$ 0.20 \end{small} \\

\bottomrule
\end{tabular}}
\caption{Performance of various unlearning methods on TOFU $1\%$ split averaged over 3 random seeds, on Llama2-7b. FQ, CI, MU, TC and FU represent Forget Quality, Cleanness Indistinguishability, Model Utility, Text Cleanness and Forget Utility, respectively. `Finetune' denotes the finetuned model on the TOFU that has yet to undergo unlearning, while `Retain' refers to the model trained solely on the retain set. 
An upward arrow ($\uparrow$) indicates that a higher value is preferable. 
The best results are highlighted in \textbf{bold}, except for MU, where bolding indicates performance on par with ``finetune''.}
\label{tab:forget01_full}
\end{table*}
% \usepackage{booktabs}
\begin{table*}[ht]
\centering
\scalebox{0.8}{
\begin{tabular}{@{}cccccc@{}}
\toprule
                    % & \multicolumn{5}{c}{TOFU 10\%}                                                                        \\ \cmidrule(lr){2-6}
\multirow{2}{*}{Method}              & \multicolumn{2}{c}{Forgetting}              & \multicolumn{3}{c}{Utility}                                \\ \cmidrule(lr){2-3}  \cmidrule(lr){4-6}
                    & FQ ($\uparrow$) & \gibfq{} ($\uparrow$) & MU ($\uparrow$) & \gib{} ($\uparrow$) & \cs{} ($\uparrow$) \\ \midrule
Finetune   & 4.3e-21          & 9.12e-4                      & 0.58            & 0.92                 & 0.98               \\
Retain     & 1.0             & 1.0                   & 0.58            & 0.95                & 0.99               \\ \midrule

GradAsc & 6.8e-07 \begin{small} $\pm$ 1.9e-07 \end{small} & 8.9e-177 \begin{small} $\pm$ 0.00 \end{small} & 0.00 \begin{small} $\pm$ 0.00 \end{small} & 0.11 \begin{small} $\pm$ 0.02 \end{small} & 0.00 \begin{small} $\pm$ 0.00 \end{small} \\
GradDiff & 1.6e-13 \begin{small} $\pm$ 8.7e-14 \end{small} & 4.0e-162 \begin{small} $\pm$ 3.1e-162 \end{small} & \textbf{0.60} \begin{small} $\pm$ 1.7e-03 \end{small} & 0.03 \begin{small} $\pm$ 1.7e-03 \end{small} & 8.9e-03 \begin{small} $\pm$ 1.9e-03 \end{small} \\
NPO & 0.12 \begin{small} $\pm$ 0.05 \end{small} & 3.0e-49 \begin{small} $\pm$ 5.1e-49 \end{small} & 0.58 \begin{small} $\pm$ 5.2e-03 \end{small} & 0.41 \begin{small} $\pm$ 0.23 \end{small} & 0.26 \begin{small} $\pm$ 0.06 \end{small} \\
IdkPO & 9.0e-03 \begin{small} $\pm$ 0.01 \end{small} & 2.5e-56 \begin{small} $\pm$ 2.2e-56 \end{small} & 0.56 \begin{small} $\pm$ 4.0e-03 \end{small} & 0.46 \begin{small} $\pm$ 0.02 \end{small} & 0.63 \begin{small} $\pm$ 0.14 \end{small} \\
AltPO & \textbf{0.51} \begin{small} $\pm$ 0.31 \end{small} & \textbf{2.8e-05} \begin{small} $\pm$ 4.4e-05 \end{small} & 0.56 \begin{small} $\pm$ 0.01 \end{small} & \textbf{0.89}\begin{small} $\pm$ 0.02 \end{small} & \textbf{0.86} \begin{small} $\pm$ 0.04 \end{small} \\

\bottomrule
\end{tabular}
}
\caption{Performance of various unlearning methods on TOFU $10\%$ split averaged over 3 random seeds, on Llama3.2-3B-Instruct. FQ, CI, MU, TC and FU represent Forget Quality, Cleanness Indistinguishability, Model Utility, Text Cleanness and Forget Utility, respectively. `Finetune' denotes the finetuned model on the TOFU that has yet to undergo unlearning, while `Retain' refers to the model trained solely on the retain set. 
An upward arrow ($\uparrow$) indicates that a higher value is preferable. 
The best results are highlighted in \textbf{bold}, except for MU, where bolding indicates performance on par with ``finetune''.}
\label{tab:forget10_full_llama3_2}
\end{table*}

\begin{figure}[!htp] % 'h' means here
    \centering
    \includegraphics[width=0.45\textwidth]{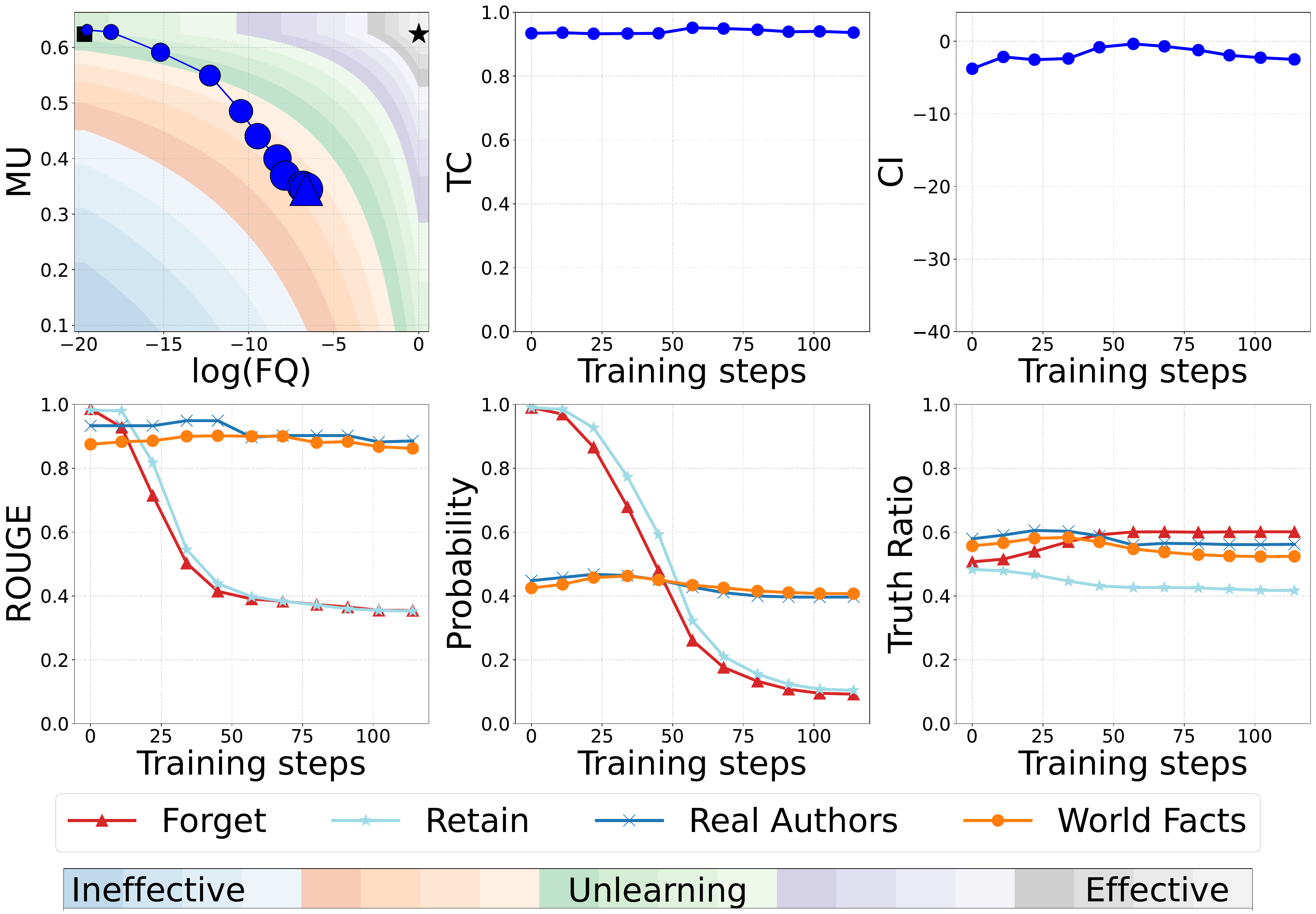} % Adjust width as needed
    % \caption{Trajectory of all evaluation metrics using GradAsc}
    \caption{GradAsc for TOFU 10\% on Llama2-7b.}
\label{plot_grad_asc_10}  
\end{figure}

\begin{figure}[!htp] % 'h' means here
    \centering
    \includegraphics[width=0.45\textwidth]{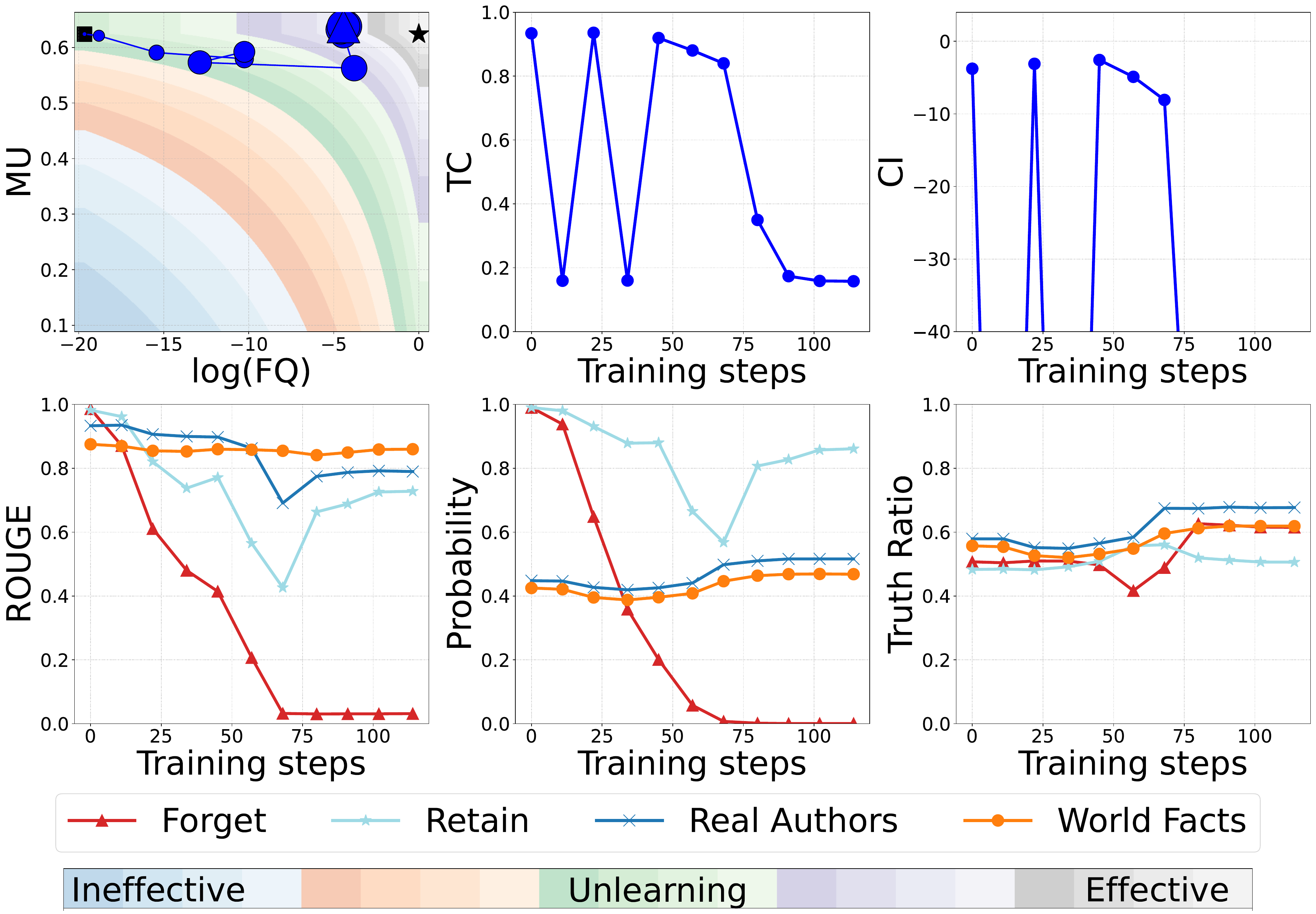} % Adjust width as needed
    % \caption{Trajectory of all evaluation metrics using GradDiff}
    \caption{GradDiff for TOFU 10\% on Llama2-7b.}
\label{plot_grad_diff_10}       
\end{figure}

\begin{figure}[!htp] % 'h' means here
    \centering
    \includegraphics[width=0.45\textwidth]{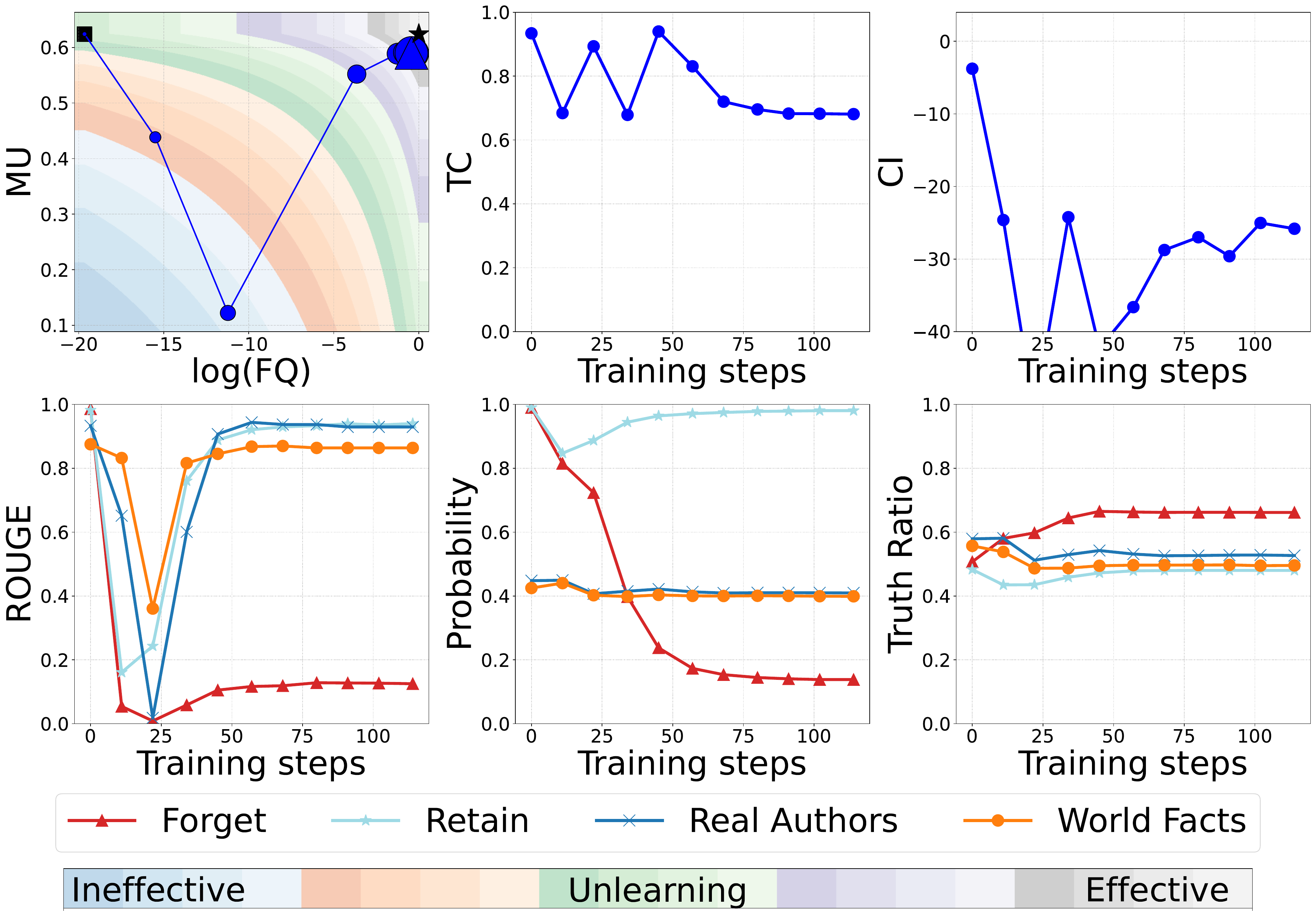} % Adjust width as needed
    % \caption{Trajectory of all evaluation metrics using IdkPO}
    \caption{IdkPO for TOFU 10\% on Llama2-7b.}
 \label{plot_idkpo_10}   
\end{figure}

\begin{figure}[!htp] % 'h' means here
    \centering
    \includegraphics[width=0.45\textwidth]{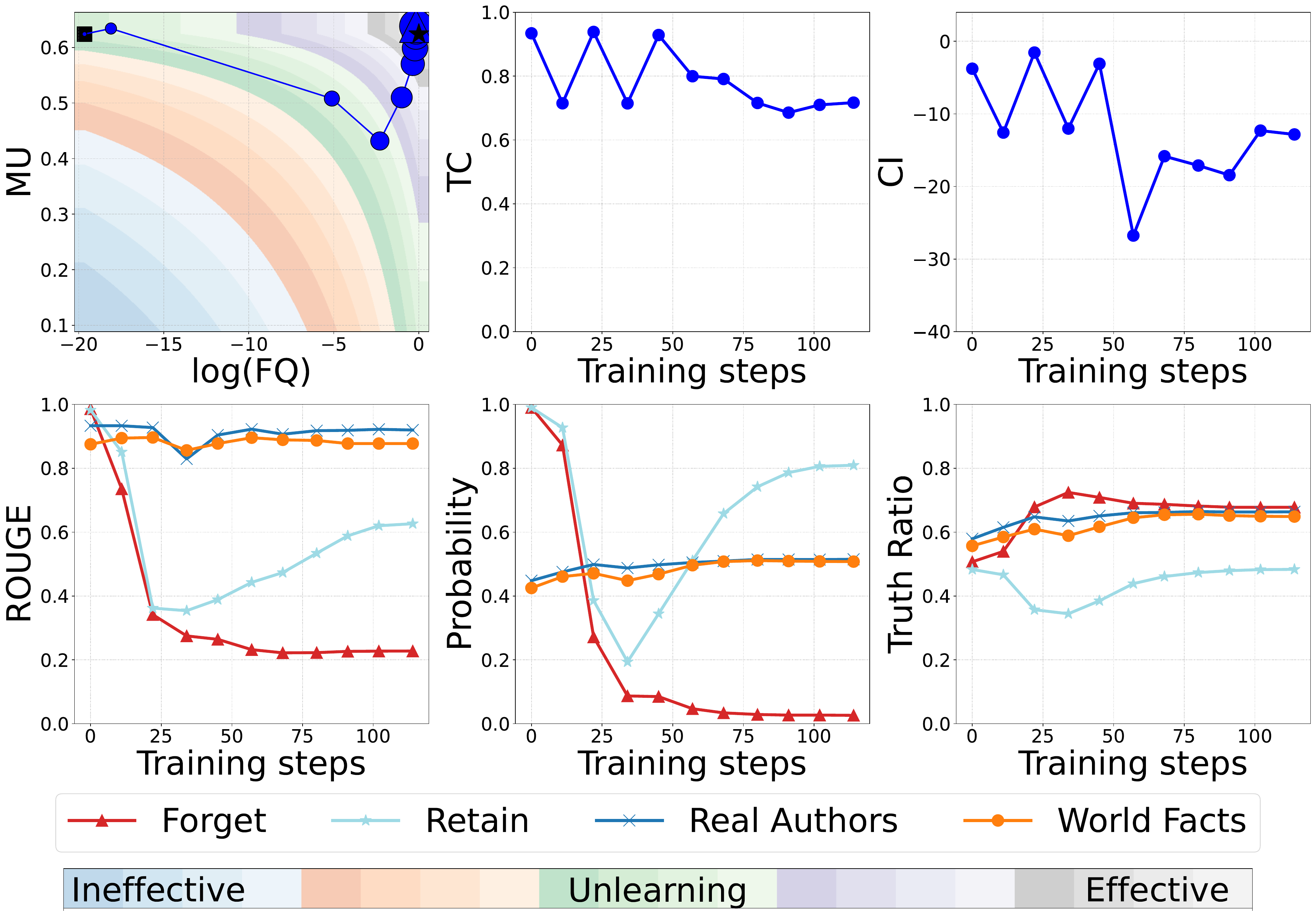} % Adjust width as needed
    % \caption{Trajectory of all evaluation metrics using NPO}
    \caption{NPO for TOFU 10\% on Llama2-7b.}
  \label{plot_npo_10}  
\end{figure}

\begin{figure}[!htp] % 'h' means here
    \centering
    \includegraphics[width=0.45\textwidth]{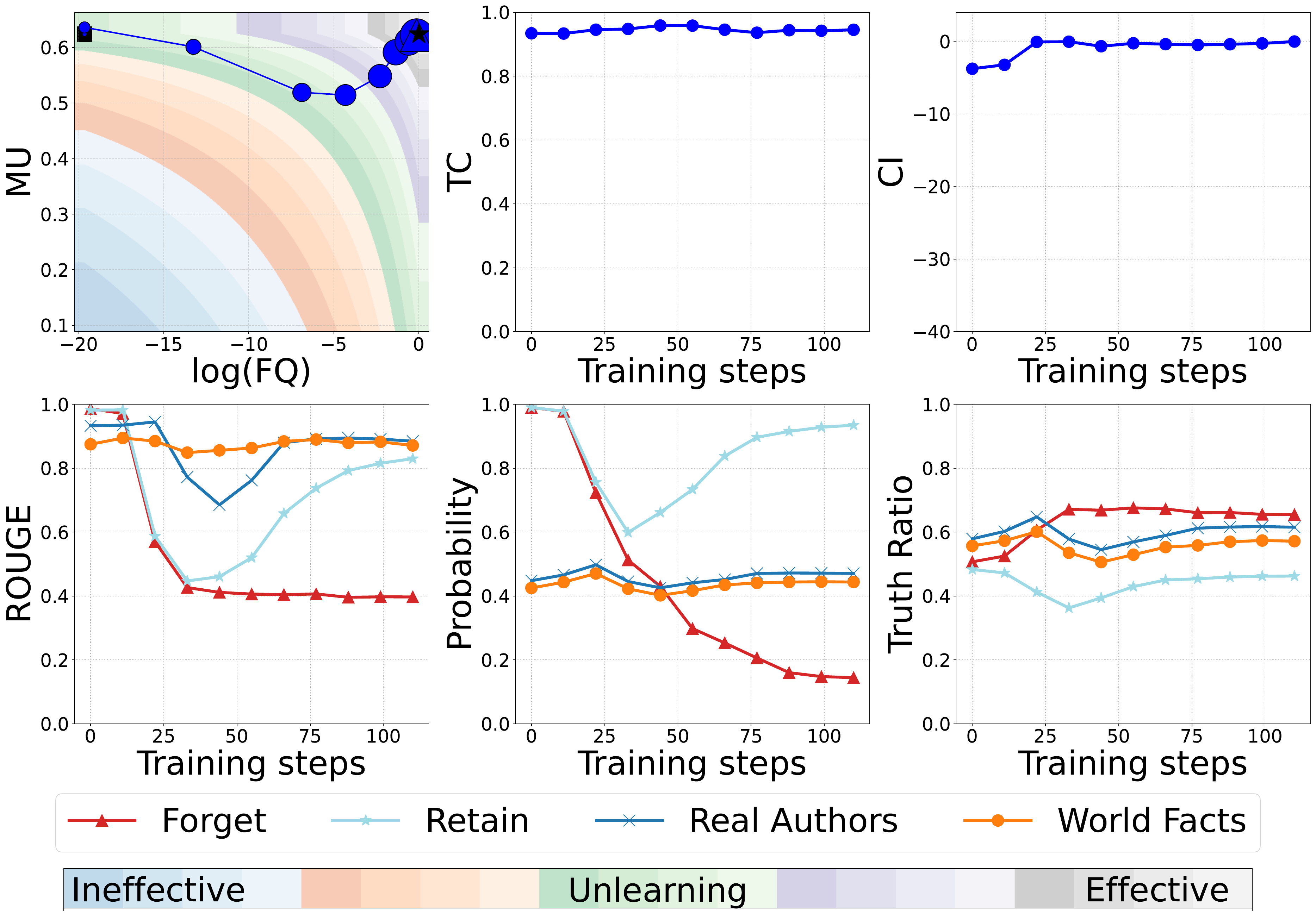} % Adjust width as needed
    % \caption{Trajectory of all evaluation metrics using AltPO}
    \caption{AltPO for TOFU 10\% on Llama2-7b.}
  \label{plot_altpo_10}
\end{figure}

\begin{figure}[!htp] % 'h' means here
    \centering
    \includegraphics[width=0.45\textwidth]{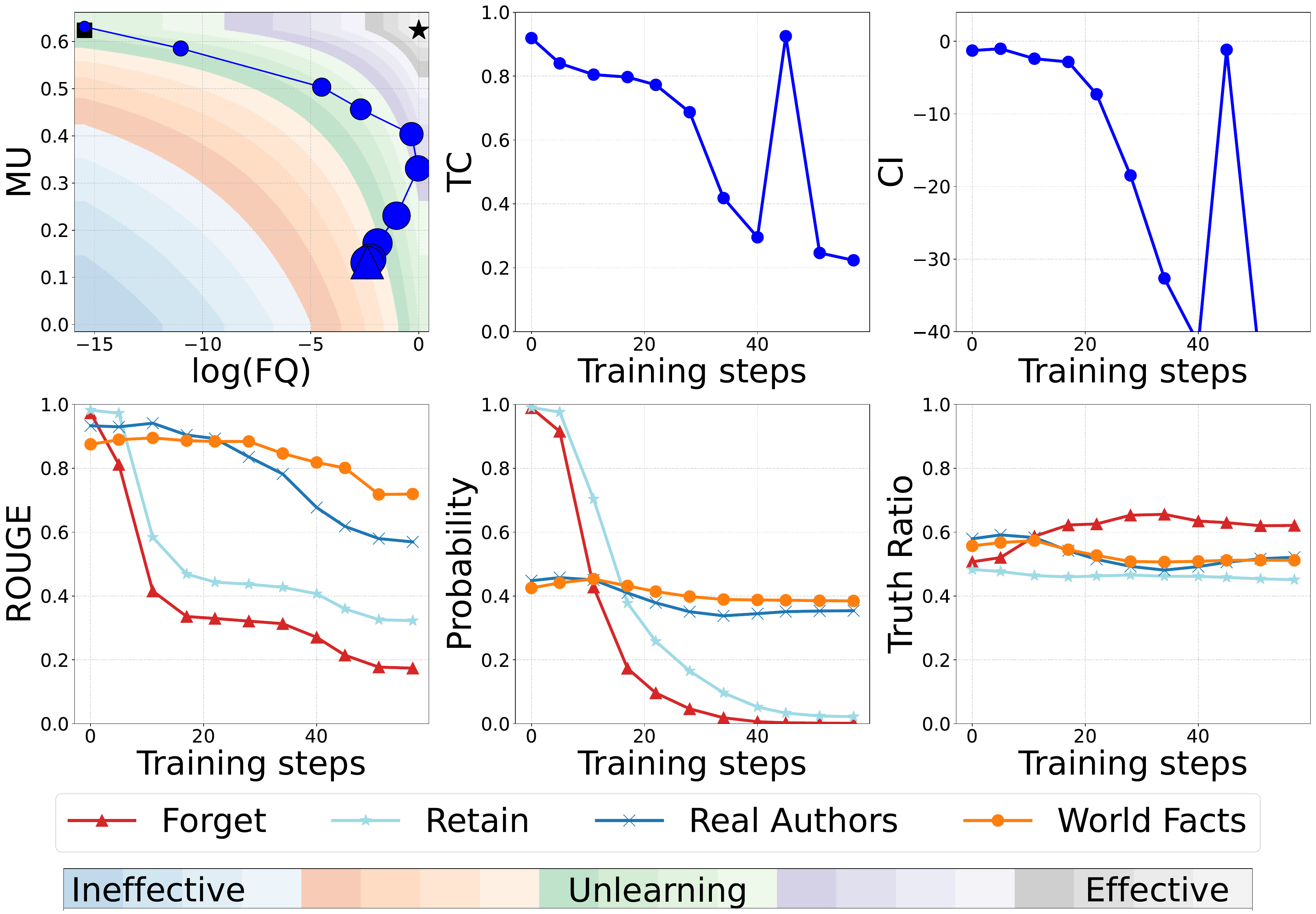} % Adjust width as needed
    % \caption{Trajectory of all evaluation metrics using GradAsc}
    \caption{GradAsc for TOFU 5\% on Llama2-7b.}
  \label{plot_5_gradasc}
\end{figure}

\begin{figure}[!htp] % 'h' means here
    \centering
    \includegraphics[width=0.45\textwidth]{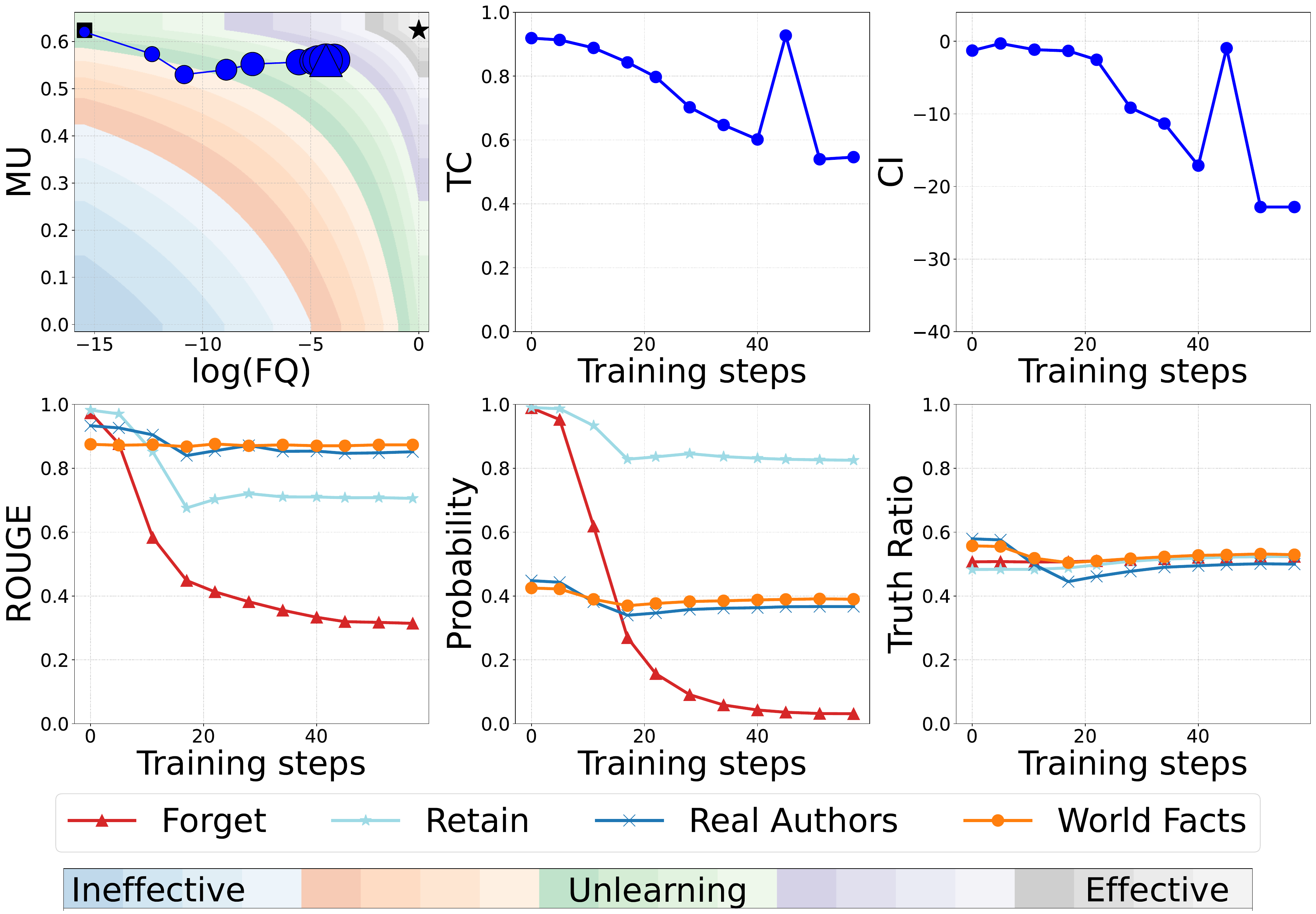} % Adjust width as needed
    % \caption{Trajectory of all evaluation metrics using GradDiff}
    \caption{GradDiff for TOFU 5\% on Llama2-7b.}
  \label{plot_5_graddiff}
\end{figure}

\begin{figure}[!htp] % 'h' means here
    \centering
    \includegraphics[width=0.45\textwidth]{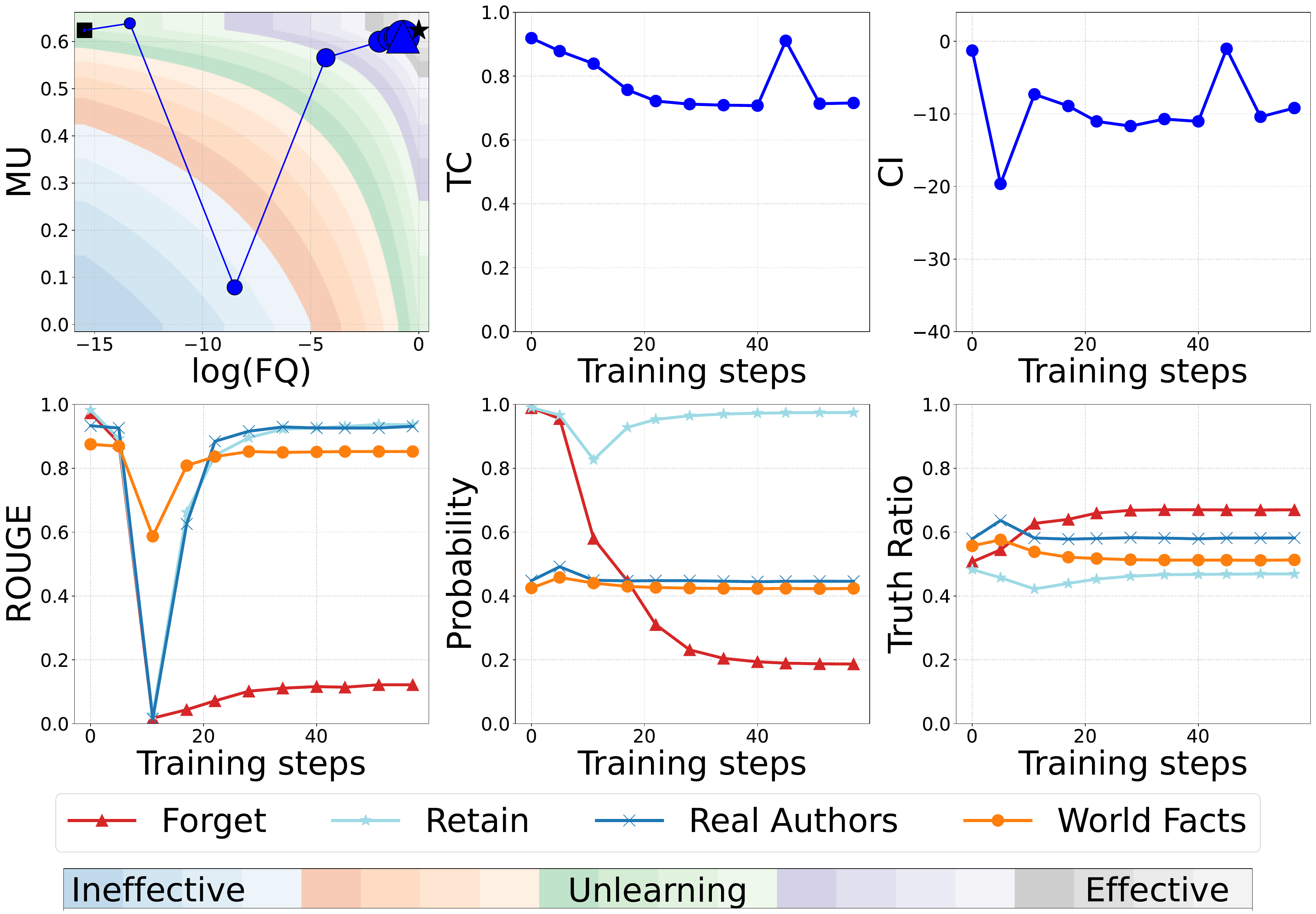} % Adjust width as needed
    % \caption{Trajectory of all evaluation metrics using IdkPO}
    \caption{IdkPO for TOFU 5\% on Llama2-7b.}
  \label{plot_5_idkpo}
\end{figure}

\begin{figure}[!htp] % 'h' means here
    \centering
    \includegraphics[width=0.45\textwidth]{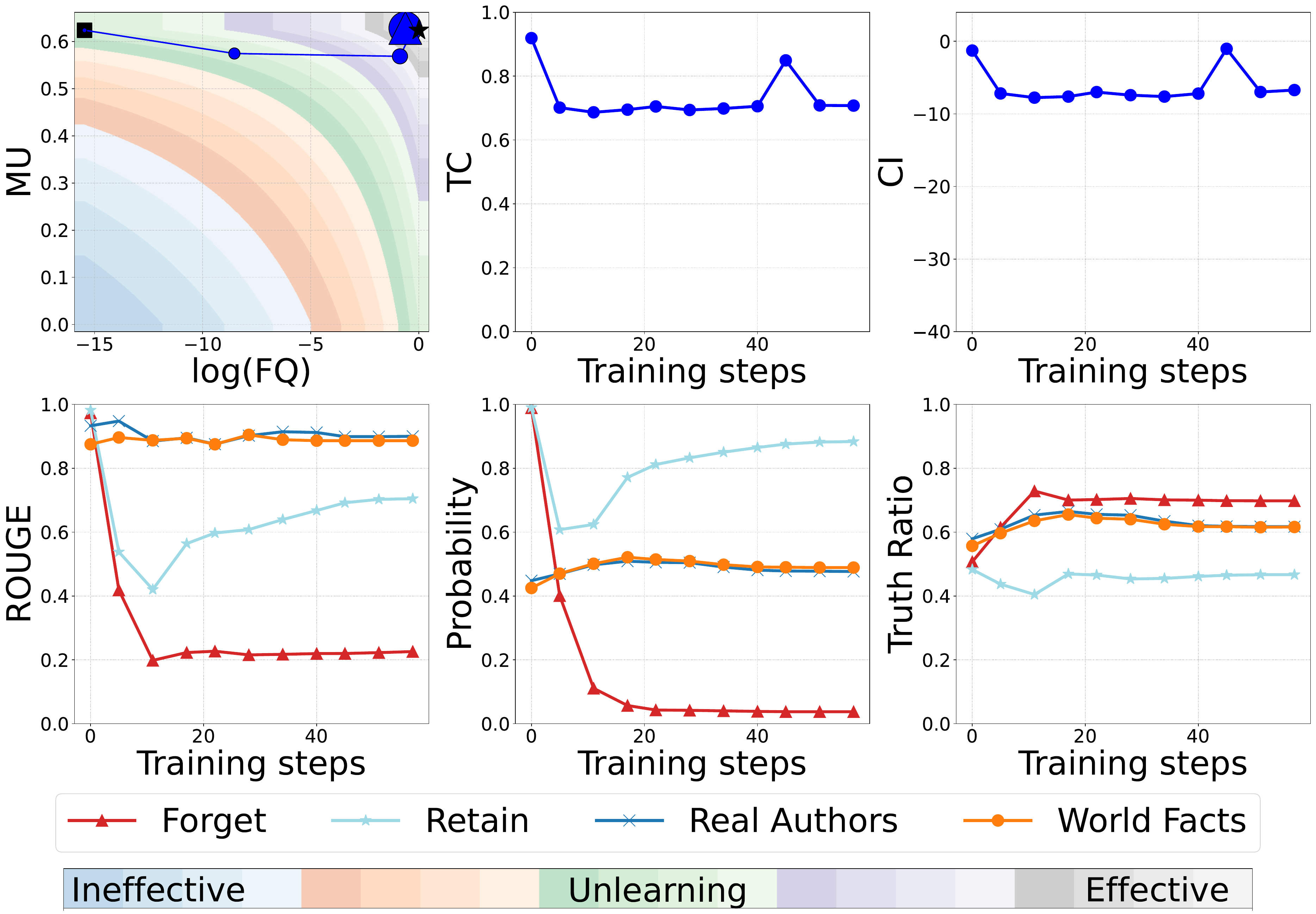} % Adjust width as needed
    % \caption{Trajectory of all evaluation metrics using NPO}
    \caption{NPO for TOFU 5\% on Llama2-7b.}
  \label{plot_5_npo}
\end{figure}

\begin{figure}[!htp] % 'h' means here
    \centering
    \includegraphics[width=0.45\textwidth]{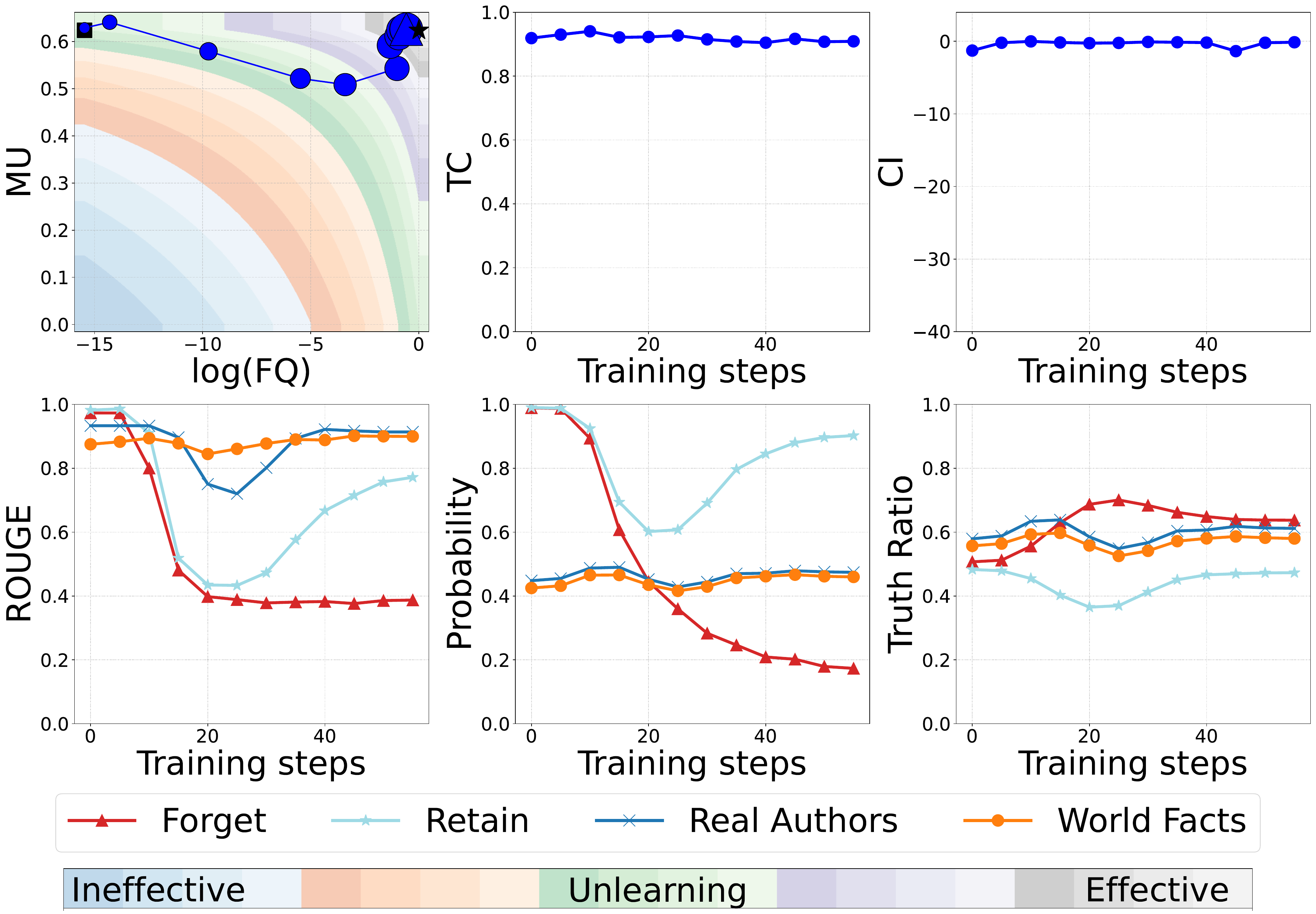} % Adjust width as needed
    % \caption{Trajectory of all evaluation metrics using AltPO}
    \caption{AltPO for TOFU 5\% on Llama2-7b.}
  \label{plot_5_altpo}
\end{figure}

\begin{figure}[!htp] % 'h' means here
    \centering
    \includegraphics[width=0.45\textwidth]{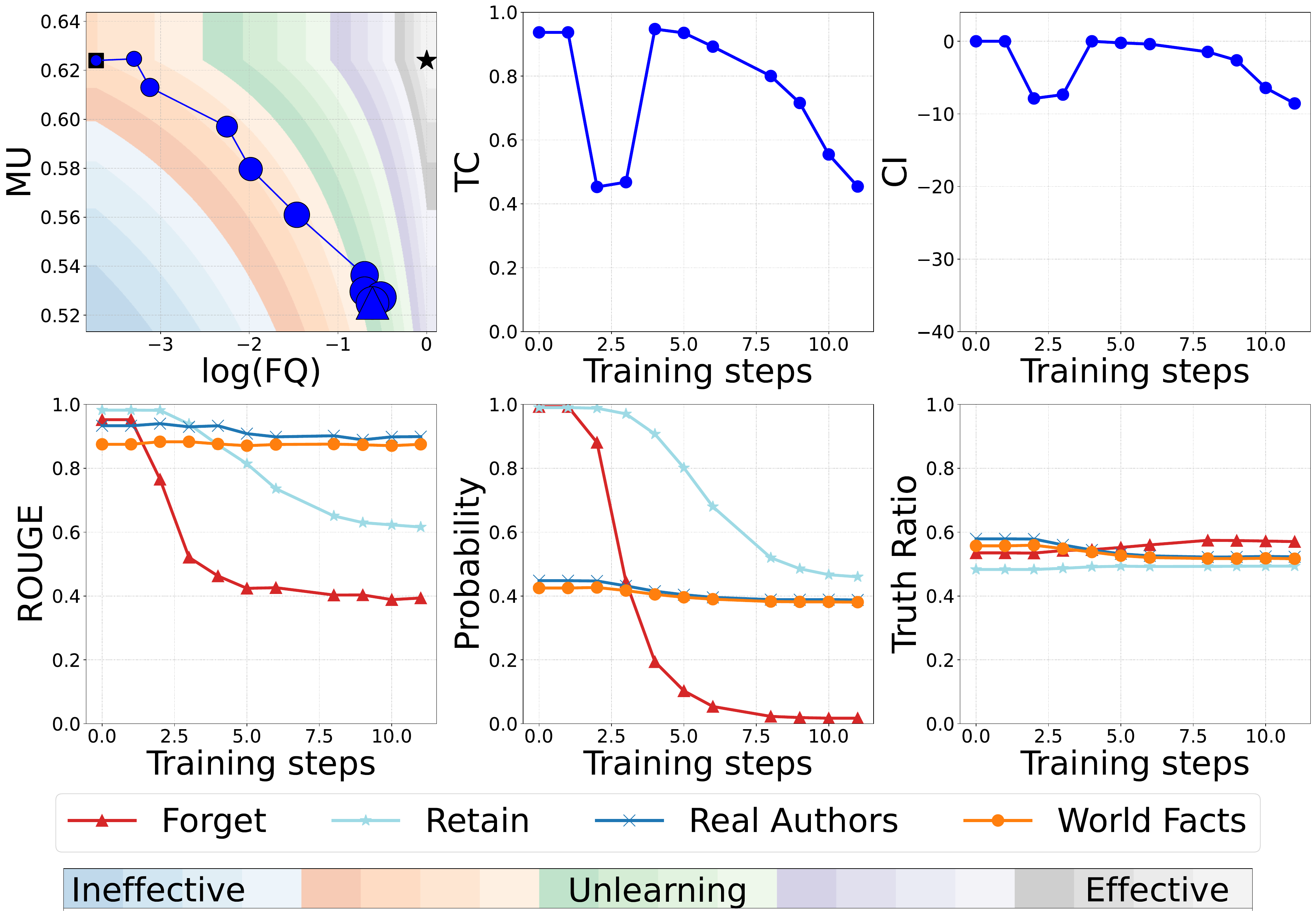} % Adjust width as needed
    % \caption{Trajectory of all evaluation metrics using GradAsc}
    \caption{GradAsc for TOFU 1\% on Llama2-7b.}
  \label{plot_1_grad_asc}
\end{figure}

\begin{figure}[!htp] % 'h' means here
    \centering
    \includegraphics[width=0.45\textwidth]{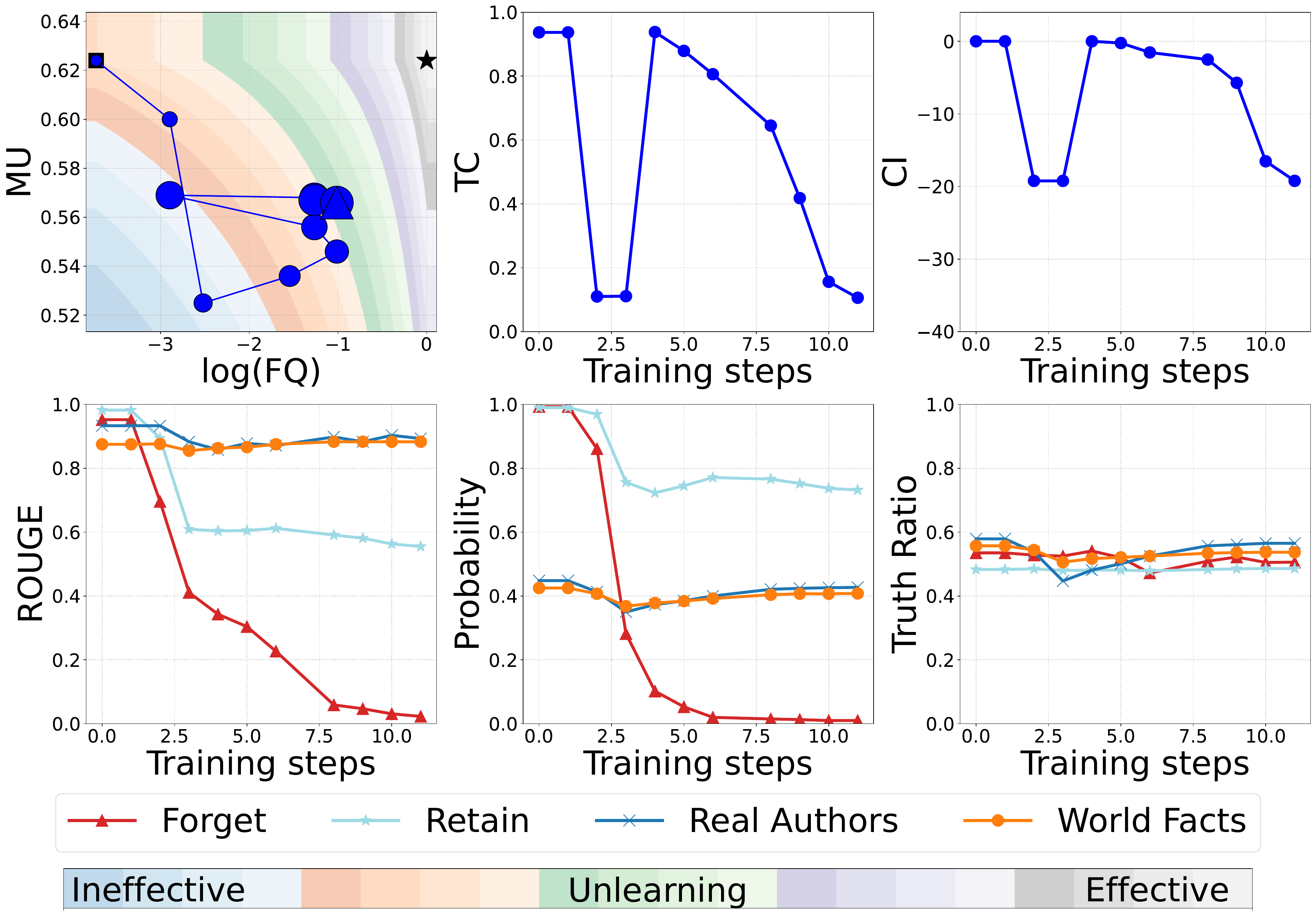} % Adjust width as needed
    % \caption{Trajectory of all evaluation metrics using GradDiff}
    \caption{GradDiff for TOFU 1\% on Llama2-7b.}
  \label{plot_1_grad_diff}
\end{figure}

\begin{figure}[!htp] % 'h' means here
    \centering
    \includegraphics[width=0.45\textwidth]{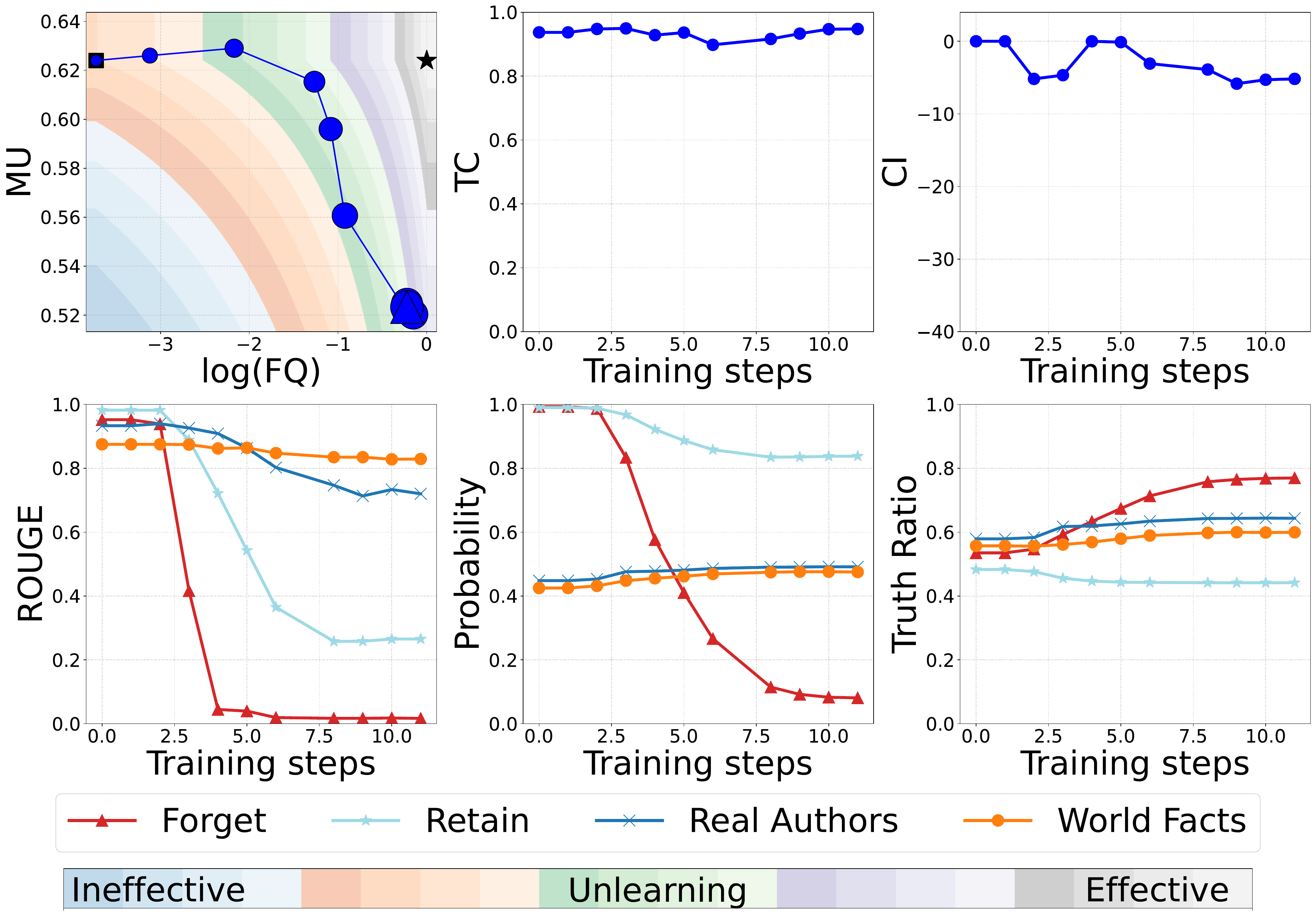} % Adjust width as needed
    % \caption{Trajectory of all evaluation metrics using IdkPO}
    \caption{IdkPO for TOFU 1\% on Llama2-7b.}
  \label{plot_1_idkpo}
\end{figure}

\begin{figure}[!htp] % 'h' means here
    \centering
    \includegraphics[width=0.45\textwidth]{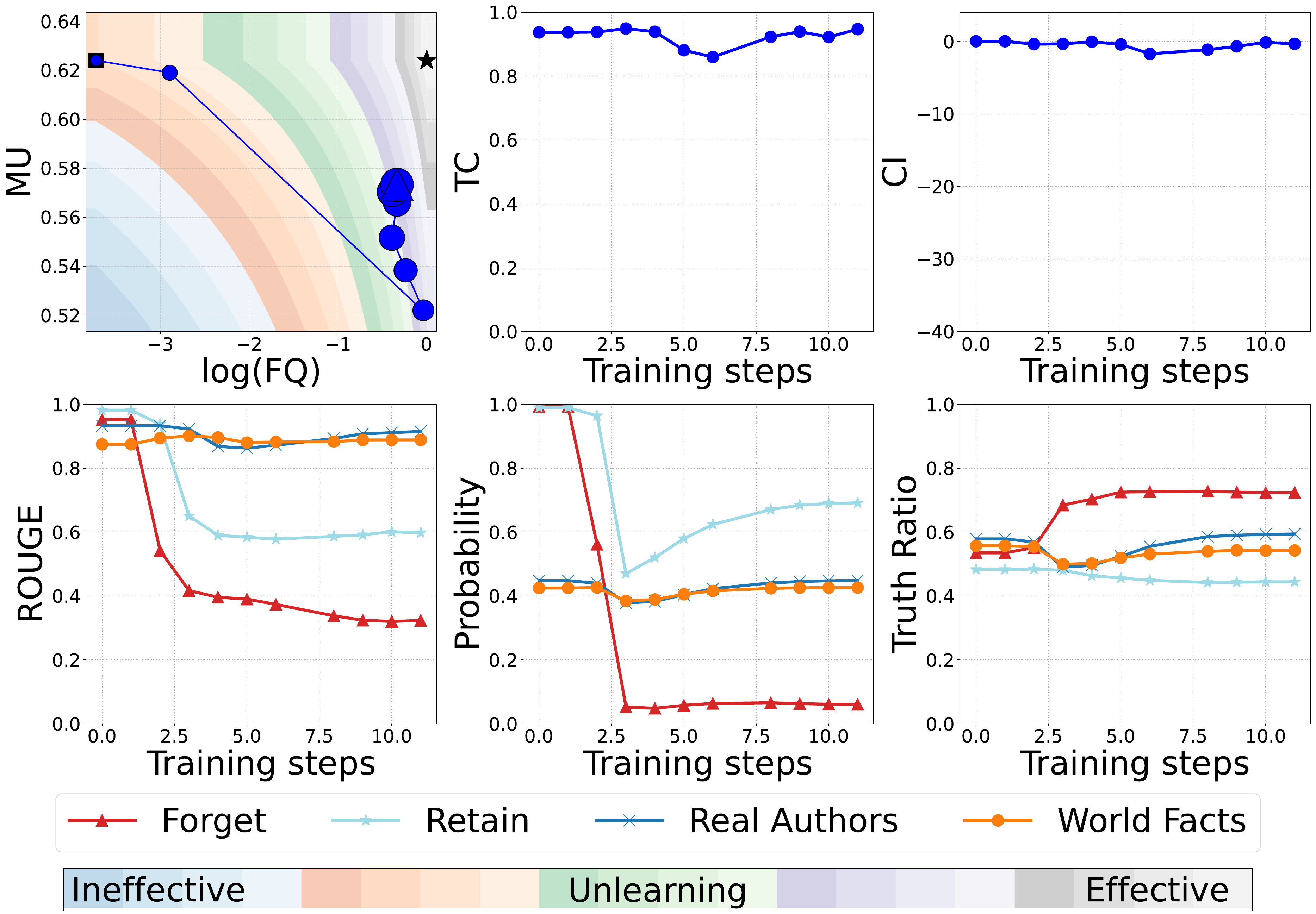} % Adjust width as needed
    % \caption{Trajectory of all evaluation metrics using NPO}
    \caption{NPO for TOFU 1\% on Llama2-7b.}
  \label{plot_1_npo}
\end{figure}

\begin{figure}[!htp] % 'h' means here
    \centering
    \includegraphics[width=0.45\textwidth]{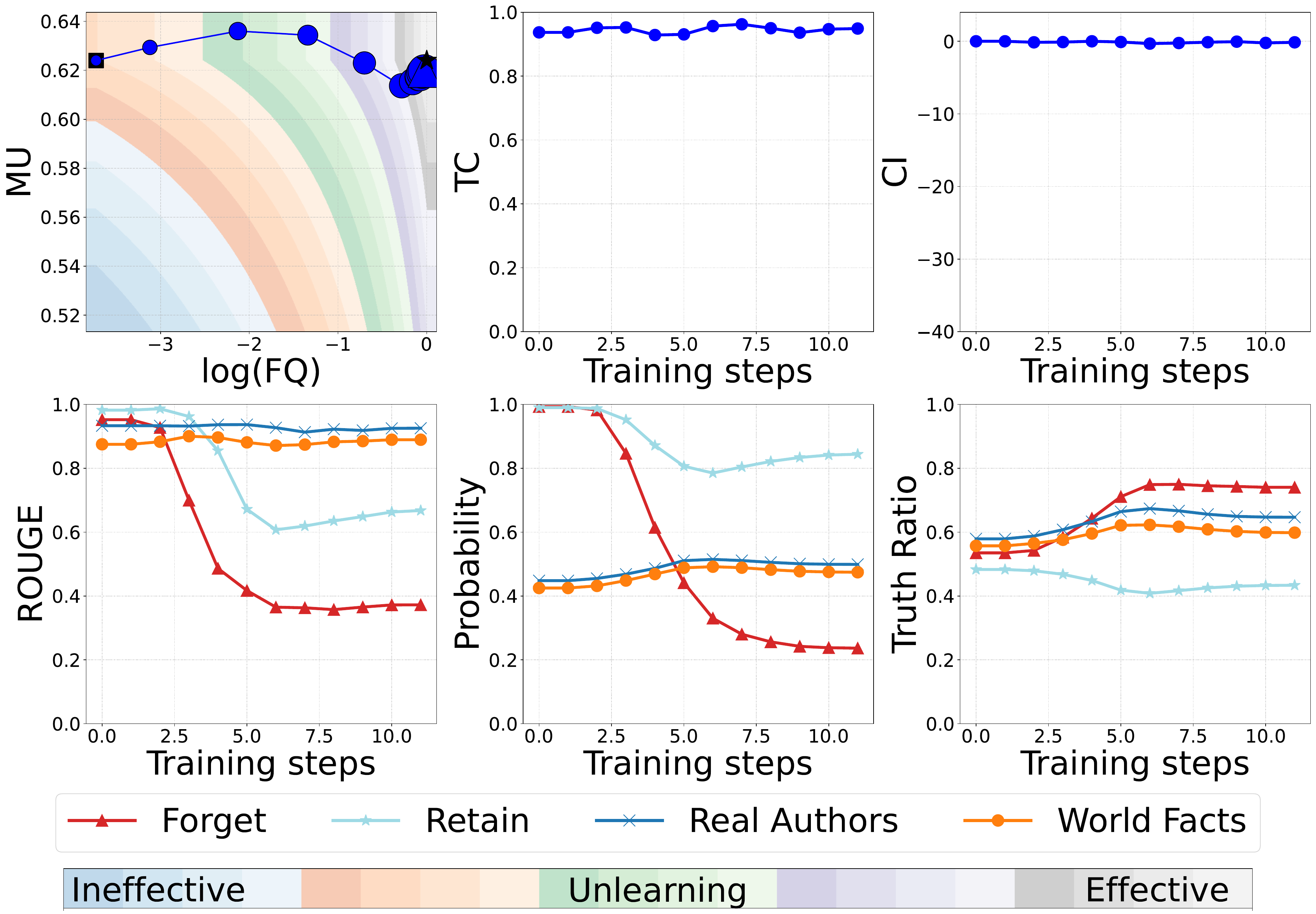} % Adjust width as needed
    % \caption{Trajectory of all evaluation metrics using AltPO}
    \caption{AltPO for TOFU 1\% on Llama2-7b.}
  \label{plot_1_altpo}
\end{figure}

The trajectories of \gibfq{} and \gib{} over the training steps are shown in \Cref{fig:CI,fig:TC}. We also provide trajectory plots for all forget splits on  Llama2, showing the variations in individual evaluation metrics, including more fine-grained metrics from TOFU in \Crefrange{plot_grad_asc_10}{plot_1_altpo}.
\clearpage

\vspace{-10pt}
        
\vspace{-10pt}

\vspace{-10pt}

\newpage

\vspace{-10pt}

\vspace{-10pt}

\vspace{-10pt}

\newpage

\vspace{-10pt}

\vspace{-10pt}

\vspace{-10pt}

\newpage

\vspace{-10pt}

\vspace{-10pt}

\newpage
\begin{table*}[t]
\centering
\scalebox{0.8}{
\begin{tabular}{ccccccc}\toprule
\multirow{2}{*}{Method} & \multirow{2}{*}{Forget Loss} & \multicolumn{2}{c}{Forgetting} & \multicolumn{3}{c}{Utility} \\ 
\cmidrule(lr){3-4} \cmidrule(lr){5-7}
                       &                         & FQ ($\uparrow$)  & CI ($\uparrow$)  & MU ($\uparrow$) & \gib{} ($\uparrow$)  & \cs{} ($\uparrow$) \\ \midrule
\subnll{}            & $\mathbb{E}_i[\text{NLL}(y^i_a | x_f)]$              & 2.2e-15      & 0.12       & \textbf{0.63 }    &0.94      & 0.97              \\ 
\ppo{}                    & $\mathbb{E}_i[\Lb_{\text{DPO}}(y^i_a, * \mid x_f)]$                 & 2.5e-12          & 0.13             & 0.61      & 0.94             & \textbf{0.98}      \\ 

\hline
\subdiff{} ($M=1$)            & $\mathbb{E}_i[\text{NLL}(y^i_a \mid x_f) - \text{NLL}(y_f \mid x_f)]$                &  1.2e-5      & 9.6e-7       & 0.55          & 0.90       & 0.90              \\
\subdiff{}           & $\mathbb{E}_i[\text{NLL}(y^i_a \mid x_f) - \text{NLL}(y_f \mid x_f)]$                & 0.14            & 0.0             & 0.61        & 0.50             & 0.48               \\ 
\hline

\methodname{} ($M=1$) & $\mathbb{E}_i[\Lb_{\text{DPO}}(y^{i}_{a}, y_f | x_f)]$ & 0.06   & 0.15     & \textbf{0.62}           & 0.93       & 0.88          \\
\methodname{}       & $\mathbb{E}_i[\Lb_{\text{DPO}}(y^{i}_{a}, y_f | x_f)]$              & \textbf{0.74}             & 0.92             & \textbf{0.62}    & 0.95        & 0.86               \\ 
\hline
$\methodname_{\text{base}}$ ($M=1$) & $\mathbb{E}_i[\Lb_{\text{DPO}}(y^{i}_{a}, y_f | x_f)]$          & 0.44    & 0.13    & \textbf{0.63}    & \textbf{0.96}      & 0.90               \\ 
$\methodname_{\text{base}}$ & $\mathbb{E}_i[\Lb_{\text{DPO}}(y^{i}_{a}, y_f | x_f)]$           & 0.71    & \textbf{0.96}    & \textbf{0.64}    & \textbf{0.96 }      & 0.90               \\ 
\bottomrule
\end{tabular}
}
\caption{Ablation study of various methods and their performance in terms of forgetting and utility. Results are reported on the $10\%$ split of the TOFU using Llama2. The best results are highlighted in \textbf{bold}, except for MU, where bolding indicates performance on par with ``finetune''. We set $M=5$ unless mentioned.}
\label{tab:ablations_appd}
\end{table*}

\end{document}